\documentclass{article}

\usepackage{microtype}
\usepackage{graphicx}
\usepackage{booktabs} 
\usepackage{enumitem}
\usepackage{caption}
\usepackage{subcaption}
\usepackage{tabularx}
\usepackage{booktabs}
\usepackage{multirow}
\usepackage{makecell}
\usepackage{mdframed}
\usepackage{breqn}
\usepackage[algo2e, lined, ruled, linesnumbered, commentsnumbered, longend, noend]{algorithm2e}

\usepackage[hypertexnames=false]{hyperref}

\newcommand{\alglinelabel}{%
  \addtocounter{ALC@line}{-1}
  \refstepcounter{ALC@line}
  \label
}

 \usepackage[accepted]{icml2025}

\usepackage{amsmath}
\usepackage{amssymb}
\usepackage{mathtools}
\usepackage{amsthm}

\usepackage[capitalize,noabbrev]{cleveref}

\theoremstyle{plain}
\newtheorem{theorem}{Theorem}[section]

\theoremstyle{definition}
\newtheorem{definition}[theorem]{Definition}

\theoremstyle{remark}

\newcommand{\dpszo}[1]{\texttt{DPS-ZO#1}}
\newcommand{\dpsmozo}[1]{\texttt{DPS-MOZO#1}}
\DeclarePairedDelimiterX{\renyidivx}[2]{(}{)}{%
  #1\;\delimsize\|\;#2%
}
\newcommand{\renyidiv}[1]{D_{#1}\renyidivx}

\SetCommentSty{mycommfont}

\DeclareMathOperator*{\argmax}{arg\,max}

\usepackage[textsize=tiny]{todonotes}

\icmltitlerunning{Differentially Private In-context Learning via Sampling Few-shot Mixed with Zero-shot Outputs}

\begin{document}

\twocolumn[
\icmltitle{Differentially Private In-context Learning via \\ Sampling Few-shot Mixed with Zero-shot Outputs}



\icmlsetsymbol{equal}{*}

\begin{icmlauthorlist}
\icmlauthor{James Flemings}{usc}
\icmlauthor{Haosheng Gan}{usc}
\icmlauthor{Hongyi Li}{usc}
\icmlauthor{Meisam Razaviyayn}{usc}
\icmlauthor{Murali Annavaram}{usc}
\end{icmlauthorlist}

\icmlaffiliation{usc}{University of Southern California}

\icmlcorrespondingauthor{James Flemings}{jamesf17@usc.edu}

\icmlkeywords{Machine Learning, ICML}

\vskip 0.3in
]



\printAffiliationsAndNotice{}  

\begin{abstract}
    In-context learning (ICL) has shown promising improvement in downstream task adaptation of LLMs by augmenting prompts with relevant input-output examples (demonstrations). However, the ICL demonstrations can contain privacy-sensitive information, which can be leaked and/or regurgitated by the LLM output. Differential Privacy (DP), a widely adopted privacy safeguard, has emerged to mitigate this privacy leakage, with recent work demonstrating strong privacy-utility tradeoffs in classification tasks for ICL. However, generation tasks for ICL are challenging due to the high-dimensional output space of open-ended generation. To this end, we propose \dpsmozo{}, Differentially Private Sampling by Mixing One-shot with Zero-shot Outputs, a decoding framework that generates DP text by sampling from the product of multiple one-shot outputs mixed with a zero-shot output. This mixing effectively reduces the amount of information that can be leaked by each demonstration. By utilizing the inherent randomness in sampling from the mixed distributions, we can achieve DP without adding noise, thereby improving the privacy-utility tradeoff. Our experimental evaluations show \dpsmozo{} can achieve a strong privacy guarantee, $\epsilon=2$, with minimal utility degradation compared to non-private few-shot learning, $\textbf{0.3\%}$ ROUGE-L F1 score decrease on the SAMSum dataset with Gemma 2 2B.
\end{abstract}
\section{Introduction}


Large language models (LLMs) exhibit an In-Context Learning (ICL) ability, enabling them to adapt to downstream tasks without modifying model parameters by conditioning the model on relevant input-output examples (demonstrations) \cite{brown2020language, min2022rethinking}. This impressive ICL ability is largely attributed to 1) the semantic prior knowledge obtained through pre-training of vast public data \cite{devlin2018bert, radford2019language} and 2) the input-output mapping from the demonstrations learned by the model during inference \cite{wei2023larger}. As a result, ICL is gaining popularity as an alternative to fine-tuning, thereby avoiding the complexities involved in LLM training.

\begin{figure}[t]
    \centering 
    \includegraphics[width=\columnwidth]{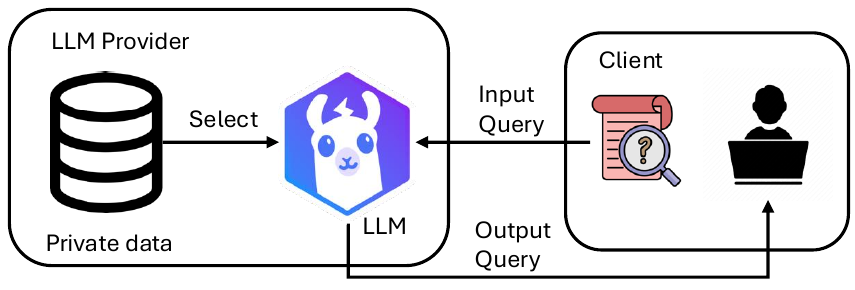}
    \caption{Problem setup for privacy-preserving in-context learning.}
    \label{fig:prob_setup}
\end{figure}

A growing concern raised is that the downstream dataset could contain sensitive information, such as Personal Identifiable Information. Although ICL does not modify an LLM's parameters to incorporate this private information, the LLM can still leak privacy by regurgitating the ICL demonstrations \cite{duan2023privacy, wang2023decodingtrust, priyanshu2023chatbots}. Consider the setup illustrated in Figure \ref{fig:prob_setup}, where a provider hosts an LLM to respond to client input queries. To enhance the quality of the responses, the provider uses demonstrations from its downstream dataset for ICL by the LLM. However, the output queries now contain information about the downstream dataset through the demonstrations, which can be leaked to the client. Hence, there is a need to develop privacy safeguards for ICL. 

\begin{figure*}[t!]
    \centering
    \begin{subfigure}[t]{0.58\textwidth}
        \includegraphics[width=1.0\columnwidth]{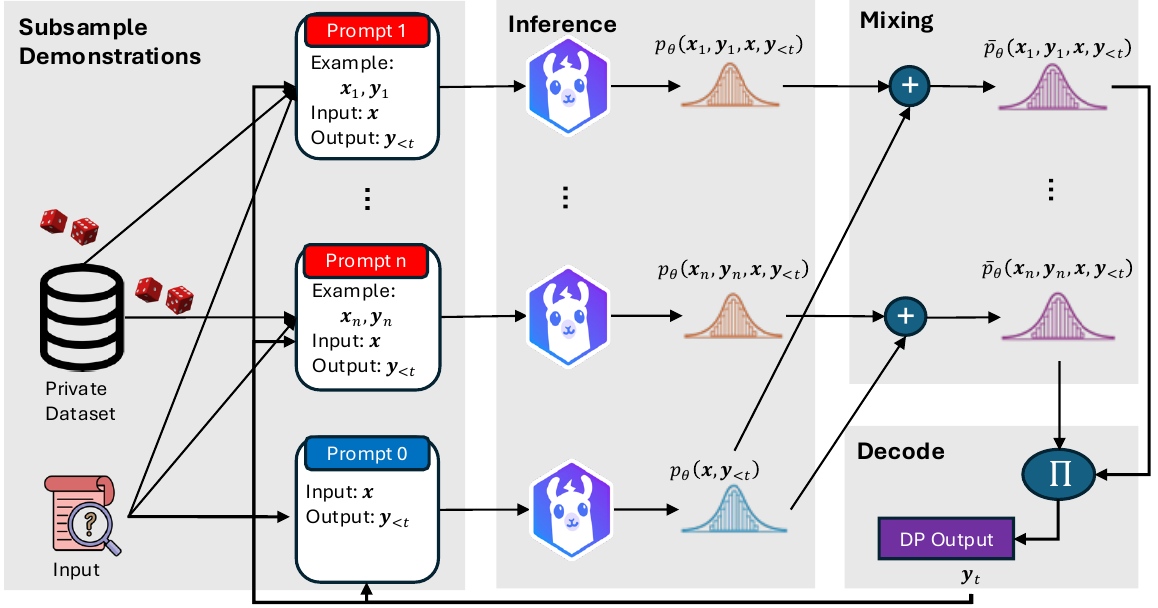}
        \caption{}
        \label{fig:dp_icl_overview}
    \end{subfigure}\hfill
    \begin{subfigure}[t]{0.4\textwidth}
        \includegraphics[width=1.0\columnwidth]{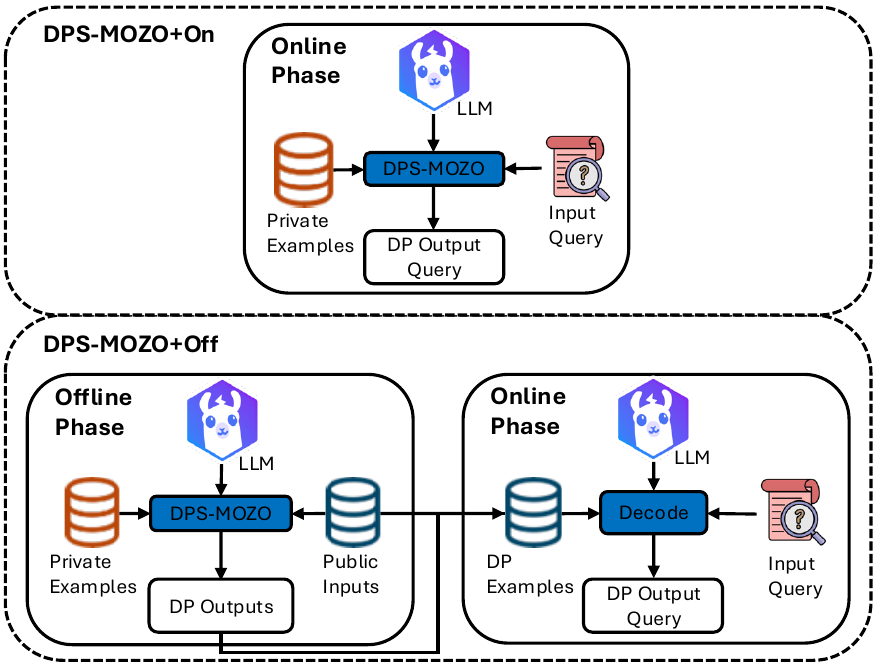}
        \caption{}
        \label{fig:dps_mozo_app}
    \end{subfigure}
    \vskip-5pt
    \caption{\textbf{(a)} A high-level overview of \dpsmozo{} broken down into four phases: (1) \textbf{Subsample Demonstrations.} First, $n_{\text{shots}}$ demonstrations are randomly selected without replacement from a private dataset, then one-shot prompts containing a demonstration, the input query, and the current output are generated. Also, a zero-shot prompt containing only the input query and the current output is generated. (2) \textbf{Inference.} An LLM performs inference on each prompt to generate $n_{\text{shots}}+1$ output distributions. (3) \textbf{Mixing.} Each one-shot output is mixed with the zero-shot output such that the resulting mixed output has a bounded divergence from the zero-shot output. (4) \textbf{Decode.} Lastly, the next token is sampled from the product of the mixed one-shot output distributions. \textbf{(b)} Two applications of \dpsmozo{} on an ICL pipeline. The first, \dpsmozo{+On}, applies \dpsmozo{} during the online phase to answer input queries. The second, \dpsmozo{+Off}, applies \dpsmozo{} during the offline phase with public inputs to generate DP outputs, both of which are to be used in the online phase.}
\end{figure*}

Differential Privacy (DP) \cite{dwork2006differential, dwork2014algorithmic} has emerged as a practically relevant solution to ameliorate privacy leakage. For classification tasks, prior work has proposed DP ICL solutions with strong privacy-utility tradeoffs \cite{wu2023privacy, tang2023privacy, duan2024flocks}. Nevertheless, achieving DP for text-generation tasks is inherently more challenging due to the high-dimensional output space of each LLM response. As an example, an LLM with a vocab size of $50,000$ and a maximum generation length of $100$ can produce $50,000^{100}$ possible different generations. \citet{wu2023privacy} overcame this by performing DP once on the entire LLM response rather than for every token in the response. However, their solution requires upwards of 100x more inferences than non-private ICL, and the final response relies heavily on the zero-shot capabilities of LLMs.

To address the privacy concerns for ICL on generation tasks with the above-mentioned limitations, we present \dpsmozo{}, Differentially Private Sampling by Mixing One-shot with Zero-shot Ouputs, a simple but effective DP decoding framework shown in Figure \ref{fig:dp_icl_overview}. The key idea of \dpsmozo{} is to bound multiple one-shot output probability distributions, the next token output distribution conditioned on an input-output example, by mixing it with the zero-shot output, the next token output that is not conditioned on any input-output examples and represents an LLM's parametric knowledge. We show that sampling from this mixed distribution can achieve DP for ICL demonstrations. Our work is broadly aligned with prior work that has shown that sampling from a bounded distribution intrinsically provides privacy \cite{wang2015privacy, husain2020local, flemings2024differentially}. 

We propose two DP ICL solutions by applying \dpsmozo{} in two different ICL settings, offline and online, to achieve DP query outputs, as shown in Figure \ref{fig:dps_mozo_app}. In the first setting, \dpsmozo{+On}, \dpsmozo{} is applied during the online phase (inference time) to generate DP query outputs for every input query received. In the second, \dpsmozo{+Off}, \dpsmozo{} is applied to public example inputs during the offline phase to generate DP example outputs as few-shot demonstrations, which will be used during the online phase. 

\textbf{A summary of our contributions} is the following:
\begin{itemize}[leftmargin=*, nosep]
    \item We introduce \dpsmozo{}, a private decoding framework that mixes multiple one-shot output distributions with zero-shot output distributions and then samples from it. This approach avoids additive noise, commonly found in DP solutions, by relying upon the inherent stochasticity of LLMs to achieve better generation quality while still achieving DP for ICL demonstrations. Moreover, \dpsmozo{} does not depend on a large ensemble of demonstrations found in current DP ICL solutions.
    \item We demonstrate that we can apply \dpsmozo{} in two different settings of the ICL pipeline to achieve DP responses: 1) the online setting, termed as \dpsmozo{+On}, which uses \dpsmozo{} to answer input queries directly and achieve DP; 2) the offline setting, \dpsmozo{+Off}, which uses \dpsmozo{} to generate DP few-shots offline that will be used for ICL in the online phase to answer input queries.
    \item Our experimental results demonstrate that we can achieve strong privacy guarantees with minimal utility degradation. In particular, we show that our framework can achieve $\epsilon=2$ while only losing $\mathbf{0.3}\%$ ROGUE-L compared to non-private few-shot ICL for the SAMSum dataset with Gemma 2 2B. 
    \item We perform extensive ablation studies on \dpsmozo{} to demonstrate its robustness to different hyperparameter values, and we conduct Membership Inference Attacks on \dpsmozo{+On} to evaluate its empirical privacy. 
\end{itemize}
\section{Related Works}\label{sec:rel_works}
Many of the prior works that have tackled differentially private in-context learning have followed the subsample-and-aggregate paradigm \cite{nissim2007smooth}. This involves subsampling a subset of the downstream dataset, partitioning the subset into smaller pairwise disjoint subsets, performing inference on each subset, and then privately aggregating the ensemble of outputs such that the final result is differentially private. Each of the prior works differs in the private aggregation aspect and the variation of the problem setup. \citet{duan2024flocks} intimately follows the PATE framework \cite{papernot2016semi, papernot2018scalable} by generating noisy labels using unlabeled public data and the argmax voting count histograms from the ensemble. \citet{tang2023privacy} generates differentially private few-shot synthetic examples using the ensemble and class labels for ICL in the online phase. 

Both \citet{duan2024flocks} and \citet{tang2023privacy} focus only on classification tasks, while \citet{wu2023privacy} extends differentially private in-context learning to summarization tasks. However, a weakness of \citet{wu2023privacy} is that every inference uses an ensemble of prompts, each of which contains multiple examples. And since their method is intended for the online inference phase, the computational cost can be quite high. For example, the ensemble size and number of shots used in their experimental setup are 100 and 4, respectively, adding an extra 100x inference cost over non-private ICL. Moreover, the final query output is still the zero-shot output of the LLM. 

For other closely related works, \citet{hanke2024open} demonstrated that local open LLMs yield higher performance and lower costs compared to closed LLMs for generation tasks. DP-OPT \cite{hong2023dp} follows a slightly different setup by performing private prompt tuning on a client's query and acts as an intermediate layer between the client and LLM provider in Figure \ref{fig:prob_setup}. AUG-PE \cite{xie2024differentially} also follows a slightly different setup where the private dataset stays on the user side rather than the LLM provider side. \dpszo{} broadly falls under private prediction methods \cite{dwork2018privacy} and builds off the work from private prediction methods for LLMs \cite{majmudar2022differentially, ginart2022submix, flemings2024differentially}.


\section{Background}
\textbf{Differential Privacy (DP)} provides a bound on the amount of information about a private dataset that the output of an algorithm can leak. We formally define this below:

\begin{definition}($(\epsilon, \delta)$-DP \cite{dwork2014algorithmic})\label{def:aprox_dp}
   Let $\epsilon \geq 0$, $\delta\in[0, 1]$. A randomized algorithm $\mathcal{A}: \mathcal{D} \rightarrow \mathcal{R}$ satisfies $(\epsilon, \delta)$-DP if for any pair of adjacent datasets $D, D' \in \mathcal{D}$ that differ in one data point and any set of subset of outputs $S \subseteq \mathcal{R}$, it holds that $\Pr[\mathcal{A}(D)\in S]  \leq e^{\epsilon} \Pr[\mathcal{A}(D')\in S] + \delta.$
\end{definition}

We use Renyi DP (RDP), another notion of DP, due to its convenient composition properties. Consequently, our framework and privacy analysis are framed in terms of RDP, but the privacy parameters in the experimental results are reported in terms of $(\epsilon, \delta)$-DP. We define Renyi Divergence.
\begin{definition}(Renyi Divergence)
    For two probability distributions $P$ and $Q$ define over $\mathcal{R}$, the Renyi divergence of order $\alpha > 1$ is 
    \begin{equation*}
        \renyidiv{\alpha}*{P}{Q} = \frac{1}{\alpha-1} \log \mathop{\mathbb{E}}_{x \sim Q}\left(\frac{P(x)}{Q(x)} \right)^{\alpha}.
    \end{equation*}
Lastly, define 
\begin{equation}\label{eq:renyi_div_sym}
    D_{\alpha}^{\leftrightarrow}(P || Q) = \max\{D_{\alpha}(P || Q), D_{\alpha}(Q || P)\}.
\end{equation}
\end{definition}

\begin{definition}($(\alpha, \epsilon(\alpha))$-RDP \cite{mironov2017renyi})
    A randomized algorithm $\mathcal{A}: \mathcal{D} \rightarrow \mathcal{R}$ is $(\epsilon(\alpha), \alpha)$-RDP if for any adjacent datasets $D, D'\in \mathcal{D}$ it holds that $\renyidiv{\alpha}*{A(D)}{A(D')} \leq \epsilon(\alpha).$
\end{definition}

\textbf{In-context Learning (ICL).} Let $D=\{(\mathbf{x}_j, \mathbf{y}_j)\}_{j=1}^n$ be a private dataset of input-output examples, $\mathbf{x}$ be an input query received from a client, and $\theta$ be the LLM parameters. Define $\text{logit}_{\theta}(\cdot)$ to be the logits output of the LLM and $p_{\theta}(\cdot) = \sigma(\text{logit}_{\theta}(\cdot))$ to be an output distribution where $\sigma$ is the softmax function. Hence, to generate an output query $\mathbf{y}$ for $\mathbf{x}$, an LLM's semantic prior knowledge $\theta$ is utilized by sampling each output token $y_t$ autoregressively from the zero-shot output distribution $p_\theta(y_t |\mathbf{x}, \mathbf{y}_{<t})$, conditioned on the input query $\mathbf{x}$ and previously generated tokens $\mathbf{y}_{<t}$. 

The output queries can be improved by including $n_{\text{shots}}$ demonstrations from $D$ for the LLM to learn the relevant input-output mapping. This is typically done by concatenating all of the demonstrations together into one prompt to obtain the output $p_\theta(y_t | \{(\mathbf{x}_j, \mathbf{y}_j)\}_{j=1}^{n_{\text{shots}}}, \mathbf{x}, \mathbf{y}_{<t})$, known as \textbf{Concat-based decoding} \cite{brown2020language}.  However, Concat-based decoding can be poorly calibrated, e.g. ordering of the demonstrations substantially changes the output \cite{zhao2021calibrate}, so an alternative method, \textbf{Ensemble-based decoding} \cite{min2021noisy}, obtains the output distribution conditioned on just one demonstration, then multiply the resulting outputs: $\prod_{j=1}^{n_{\text{shots}}} p_\theta(y_t | \mathbf{x}_j, \mathbf{y}_j, \mathbf{x}, \mathbf{y}_{<t})$. Hence, Ensemble-based decoding eliminates the dependency on the ordering of demonstrations. Our work will use ensemble-based decoding to generate the next token.

\textbf{Problem Setup.} Following \cref{fig:prob_setup}, the goal of our work is to design a privacy-preserving decoding algorithm for the LLM provider to generate an output query $\mathbf{y}$ for every input query it receives $\mathbf{x}\in D_{\text{test}}$ using an LLM $p_{\theta}$. In total, we will generate $n_{\text{test}}$ outputs. The provider wants to utilize a set of private input-output examples from a private dataset $D$ for ICL to improve downstream utility. However, the output query $\mathbf{y}$ must satisfy the $(\epsilon, \delta)$-DP guarantee (\cref{def:aprox_dp}) with respect to $D$. Our threat model assumes that an adversary only has block-box access to the LLM $\theta$, i.e., they can only query the model via an API, which returns just the output tokens, not the output logits. 
\section{Methodology}
We will now introduce our proposed framework. \cref{sec:dpsmozo} describes \dpsmozo{} and \cref{sec:dspmozo_app} shows how we can use \dpsmozo{} to achieve privacy-preserving ICL.

\subsection{\dpsmozo{}}\label{sec:dpsmozo}
As discussed in \cref{sec:rel_works}, previous works have followed the subsample-and-aggregate paradigm by generating an ensemble of output distributions and then privately aggregating the ensemble to achieve a differentially private output. However, a limitation of this scheme is that the dimensionality of each output distribution can be upwards of 50,000, so the private aggregation can have a deleterious effect on the generation quality. Instead, our approach achieves differential privacy by following another line of work that samples directly from output distributions \cite{wang2015privacy, husain2020local, flemings2024differentially}. In particular, we mix private and public output distributions such that the resulting mixed distributions have a bounded Renyi Divergence from the public distribution. In our case, the one-shot and zero-shot outputs are the private and public distributions, respectively. \cref{alg:dpsmozo} succinctly describes how \dpsmozo{} generates the next token $y_t$ and we go through each part in-detail below:

\begin{algorithm}[t!]
    \caption{\dpsmozo{}: Differentially Private Sampling}
    \label{alg:dpsmozo}
    \textbf{Input:} An LLM $\theta$, a set of input-output examples $D$, an input query $\mathbf{x}$, number of top public indices $k$, Target leakage $\beta$, Renyi order $\alpha$, max generation size $T_{\max}$, number of shots $n_{\text{shots}}$  \\ 
    \textbf{Output:} output query $\mathbf{y}$
    \begin{algorithmic}[1]
        \STATE $\mathbf{y} \gets []$
        \FOR{$t=1, ..., T_{\max}$}
        \STATE $S_t \gets$ Sample $n_{\text{shots}}$ without replacement from $D$ \label{line:subsamp}
        \STATE \textcolor{blue}{// Public top-k indices selected from zero-shot output}
        \STATE $K \gets$ top-k indices from $\text{logit}_{\theta}(\mathbf{x}, \mathbf{y}_{<t})$ \alglinelabel{line:top-k} 
        \STATE \textcolor{blue}{// Truncate zero-shot output with public indices}
        \STATE $\text{logit}_{\theta}(y_t | \mathbf{x}, \mathbf{y}_{<t}) \gets -\infty$ $\forall y_t \in K$ \alglinelabel{line:zero_shot}
        \FOR{$(\mathbf{x}_i, \mathbf{y}_i)\in S_t$}
            \STATE \textcolor{blue}{// Truncate one-shot output with public indices}
            \STATE $\text{logit}_{\theta}(y_t |\mathbf{x}_i, \mathbf{y}_{i}, \mathbf{x}, \mathbf{y}_{<t})= -\infty$ $\forall y_t \in K$ \alglinelabel{line:post_trunc}
            \STATE \textcolor{blue}{// Bound divergence of mixed output distribution}
            \STATE $\overline{p}_{\theta, \lambda_i}(\mathbf{x}_i, \mathbf{y}_{i}, \mathbf{x}, \mathbf{y}_{<t}) \!\gets\! \sigma(\lambda_i \text{logit}_{\theta}(\mathbf{x}_i, \mathbf{y}_{i}, \mathbf{x}, \mathbf{y}_{<t}) \newline + (1-\lambda_i)\text{logit}_{\theta}(\mathbf{x}, \mathbf{y}_{<t}))$ \alglinelabel{line:mix}
            \STATE $\textstyle \lambda_i \gets \argmax_{\lambda_i\geq 0} D_{\alpha}^{\leftrightarrow}(\overline{p}_{\theta, \lambda_i}(\mathbf{x}_i, \mathbf{y}_{i}, \mathbf{x}, \mathbf{y}_{<t}) || \newline p_{\theta}(\mathbf{x},$ $\mathbf{y}_{<t})) \leq \beta\alpha$ \alglinelabel{line:set_lambd}
        \ENDFOR
        \STATE \textcolor{blue}{// Ensemble-based decoding}
        \STATE $y_{t} \sim \prod_{i=1}^{n_{\text{shots}}} \overline{p}_{\theta, \lambda_i}(y_t |\mathbf{x}_{i}, \mathbf{y}_i, \mathbf{x}, \mathbf{y}_{<t})$ \alglinelabel{line:ensemble}
        \STATE $\mathbf{y} \gets \mathbf{y} + [y_t]$
        \IF{$y_t == \texttt{<eos>}$}{
            \STATE \textbf{Return} $\mathbf{y}$
        }
        \ENDIF
        \ENDFOR
    \STATE \textbf{Return} $\mathbf{y}$
    \end{algorithmic}
\end{algorithm}

\textbf{Privacy Amplification by Subsampling.} First, we observe that ICL only needs a very small subset of the total input-output examples $D$ to be effective. Hence, we can take advantage of this setup by randomly sampling $n_{\text{shots}}$ without replacement from $D$ (Line~\ref{line:subsamp}). The benefit of subsampling is it provides additional privacy to the overall downstream dataset $D$ (see \cref{thm:amp_subsamp}). In particular, if an algorithm is $(\epsilon, \delta)$-DP then with subsampling probability $q=n_{\text{shots}}/|D|$ it is roughly $(O(q\epsilon), q\delta)$-DP \cite{balle2018privacy}.
    
\textbf{Inference.} Next, we generate one-shot prompts that contain a subsampled demonstration $(\mathbf{x}_i, \mathbf{y}_i)\in S_t$, the input query $\mathbf{x}$, and previously generated tokens $\mathbf{y}_{<t}$. Additionally, we generate a zero-shot prompt that only contains the input query and previously generated tokens. Examples of one-shot and zero-shot prompts can be found in \cref{app:add_exp_setup}. Then, we perform inference on each prompt to obtain one-shot $\text{logit}_{\theta}(\mathbf{x}_i, \mathbf{y}_i, \mathbf{x}, \mathbf{y}_{<t})$ and a zero-shot $\text{logit}_{\theta}(\mathbf{x}, \mathbf{y}_{<t}))$.
    
\textbf{Output Space Truncation.} Additionally, we follow \citet{tian2022seqpate, tang2023privacy} by obtaining top-$k$ indices $K$ from a public distribution (in our case, this is the zero-shot output distribution, Line~\ref{line:top-k}). The public indices $K$ are used to truncate and rescale the output space of the zero-shot output (Line~\ref{line:zero_shot}) and the one-shot outputs (Line~\ref{line:post_trunc}). Selecting $K$ relies purely on the zero-shot output and, hence, does not leak information about the private dataset $D$. The benefits of truncating the output space for our use-case are two-fold: (1) truncation methods such as top-k sampling \cite{fan2018hierarchical} have been shown to substantially improve generation quality by reducing the selection of unlikely tokens; (2) the re-scaling of the zero-shot and one-shot output distributions potentially converge the two, which allows for less reliance on zero-shot capabilities (higher $\lambda_i$ in Line~\ref{line:set_lambd}).
    
\textbf{Output Space Mixing.} Then, each one-shot logit output is mixed with the zero-shot logit output by using the following:
    \begin{align}
        \textstyle \overline{p}_{\theta, \lambda_i}(\mathbf{x}_i, \mathbf{y}_i, \mathbf{x}, \mathbf{y}_{<t}) &\gets \sigma\big(\lambda_i \text{logit}_{\theta}(\mathbf{x}_i, \mathbf{y}_i, \mathbf{x}, \mathbf{y}_{<t}) \nonumber \\
        &+ (1-\lambda_i)\text{logit}_{\theta}(\mathbf{x}, \mathbf{y}_{<t})\big).\label{eq:mix} 
    \end{align}
    The formulation of \cref{eq:mix} is inspired by \citet{xu2023context} and \citet{flemings2024differentially}. The hyperparameter $\lambda_i$ controls how much privacy is potentially leaked by $(\mathbf{x}_i, \mathbf{y}_i)$. 
    \begin{itemize}[leftmargin=*, nosep]
    \item $\lambda_i=0$: the mixed output simplifies to zero-shot output, relying purely on the model's semantic prior knowledge.
    \item $0\leq\lambda_i < 1$: the mixed output is a weighted average between one-shot and zero-shot output, reducing privacy leakage of the one-shot output by placing weight on the zero-shot output.
    \item $\lambda_i=1$: the mixed output simplifies to the one-shot output, fully utilizing the input-output mapping. 
    \item $\lambda_i > 1.0$: the mixed output amplifies the difference between the one-shot and zero-shot output, effectively down-weighting the prior knowledge. This has been shown to help improve model utility \cite{xu2023context}.
    \end{itemize}
    
Following \citet{flemings2024differentially}, each $\lambda_i$ is selected by the following optimization problem:
\begin{align}
      &\textstyle \lambda_i \gets \max_{\lambda_i} \quad \lambda_i \label{eq:lambd_solver} \\
     &\textrm{s.t.} \quad D_{\alpha}^{\leftrightarrow}(\overline{p}_{\theta, \lambda_i}(\mathbf{x}_i, \mathbf{y}_i, \mathbf{x}, \mathbf{y}_{<t}) || p_{\theta}(\mathbf{x}, \mathbf{y}_{<t})) \!\leq\! \beta\alpha \nonumber
\end{align}
where $D_{\alpha}^{\leftrightarrow}(\cdot || \cdot)$ is defined in \cref{eq:renyi_div_sym}. The intuition behind \cref{eq:lambd_solver} is we are adaptively controlling the amount of influence each demonstration has on the output of $\dpsmozo{}$ given a target leakage $\beta$. Generally speaking, if $\lambda_i$ is large, then it indicates that the demonstrations do not have a substantial effect on the output of the LLM. In other words, the LLM can rely more on its parametric knowledge even when conditioned on an input-output example, so the one-shot and zero-shot outputs are similar.  
    
Stated more formally, \cref{eq:lambd_solver} finds the largest $\lambda_i$ that projects the mixed distribution (\cref{eq:mix}) onto a Renyi Divergence ball centered at the zero-shot distribution with radius $\beta\alpha$. We discuss how $\beta$ is chosen in \cref{sec:priv_anlaysis}. This is solved numerically using the bisection method from the \texttt{SciPy} library rather than solved analytically, which can result in loose, imprecise bounds. In our implementation, we limit the value to  $0 \leq \lambda_i \leq 1.5$ as \citet{xu2023context} found that $\lambda=1.5$ generally yields good results across all settings and all datasets. 

\textbf{Decoding.} Lastly, we use the resulting mixed distributions to perform Ensemble-based decoding. More precisely, we multiply them together, then sample from the resulting product (Line~\ref{line:ensemble}): $y_{t} \sim \prod_{i=1}^{n_{\text{shots}}} \overline{p}_{\theta, \lambda_i}(y_t |\mathbf{x}_{i}, \mathbf{y}_i, \mathbf{x}, \mathbf{y}_{<t}).$

\textbf{Privacy Analysis.} We provide a detailed privacy analysis of \dpsmozo{} in Appendix \ref{sec:priv_anlaysis}. At a high level, our analysis begins by showing that \dpsmozo{} achieves DP for generating the next token. Next, we demonstrate that the subsampling in Line~\ref{line:subsamp} provides privacy amplification for each token generation. Lastly, we derive the final privacy loss by composing the subsampled privacy loss across $n_{\text{test}} T_{\max}$ iterations, where $n_{\text{test}}$ is the number of input queries, then converting it from $(\epsilon(\alpha), \alpha)$-RDP to $(\epsilon, \delta)$-DP.

\subsection{Applying \dpsmozo{} for Privacy-preserving ICL}\label{sec:dspmozo_app}
Now we show two ways to use \dpsmozo{} that achieve DP ICL. We follow the same ICL setup from \citet{tang2023privacy} by utilizing two phases: (I) the offline phase, where the provider performs any necessary preprocessing; (II) the online phase, where the provider answers queries received by users. Pseudocode for both solutions can be found in \cref{alg:dpsmozo_on} and \cref{alg:dpsmozo_off} of \cref{app:add_algs}.

\textbf{\dpsmozo{+On}.} The first application of \dpsmozo{} is the most straightforward. No operations are needed during the offline phase. The provider uses \dpsmozo{}, the input query $\mathbf{x}$, and the private dataset $D$, to generate DP output queries during the online phase. However, a limitation of this approach is that the privacy guarantee $\epsilon$ decays with each additional query answered. Hence, the provider will need to replace the private dataset $D$ after exhausting the allocated number of queries.  

\textbf{\dpsmozo{+Off}.} To avoid replacing the entire private dataset, the provider can instead utilize the offline phase to generate DP few-shot demonstrations, which will be used for ICL during the online phase. We follow the PATE setup \cite{papernot2016semi} by utilizing a small set of unlabeled public inputs that has similar distribution as the private dataset. Hence, \dpsmozo{} uses the private dataset and the public inputs to generate DP outputs, which will be paired with the public inputs as ICL demonstrations. An advantage of this setup is that due to the post-processing property of DP (see \cref{thm:post-processing}), the use of DP few-shot demonstrations does not leak additional information about the private dataset. Consequently, the provider can employ any non-private decoding method on the DP few-shots without any query restrictions.

\section{Experiments}
\begin{table*}[t!]
    \caption{Evaluation on various ICL generation tasks utilizing 4 shots ($n_{\text{shots}}=4$). We compare our method against \citet{wu2023privacy} evaluated on different $\epsilon$ while setting $\delta = 1 / |D|$. The highest utility across all methods at a given $\epsilon$ is \textbf{bolded}.}
    \label{tbl:main_results}
    \begin{center}
    \resizebox{\linewidth}{!}{%
    \begin{tabular}{l|cc|cc|cc|cc}
        \toprule 
        & & & \multicolumn{2}{c}{\textbf{SAMSum}} & \multicolumn{2}{c}{\textbf{E2E}} & \multicolumn{2}{c}{\textbf{WikiLarge}} \\
        \cmidrule(lr){4-5}
        \cmidrule(lr){6-7}
        \cmidrule(lr){8-9}

        \textbf{Model} & \textbf{Privacy} & \textbf{Method} & ROUGE-L & BERTScore & ROUGE-L & BERTScore & ROUGE-L & BERTScore \\
        \midrule
        \multirow{12}{*}{LLaMA 3 3B} 
        & $\epsilon=0$ & Zero-shot & $22.13_{\pm 1.14}$ & $76.13_{\pm 0.44}$ & $31.12_{\pm 1.04}$ & $80.01_{\pm 0.46}$ & $41.04_{\pm 1.21}$ & $80.59_{\pm 0.40}$ \\
        \cmidrule(lr){2-9}
        & \multirow{3}{*}{$\epsilon=1$} 
        & \citet{wu2023privacy} & $25.42_{\pm 1.09}$ & $77.62_{\pm 0.25}$ & $40.03_{\pm 0.59}$ & $84.40_{\pm 0.11}$ & $49.52_{\pm 0.92}$ & $82.99_{\pm 0.34}$ \\
        & & \dpsmozo{+On} & $\mathbf{26.92}_{\pm 1.16}$ & $\mathbf{78.04}_{\pm 0.32}$ & $42.96_{\pm 1.24}$ & $84.73_{\pm 0.19}$ & $49.58_{\pm 1.02}$ & $83.08_{\pm 0.25}$ \\
        & & \dpsmozo{+Off} & $26.82_{\pm 0.71}$ & $77.82_{\pm 0.20}$ & $\mathbf{43.28}_{\pm 1.07}$ & $\mathbf{85.04}_{\pm 0.36}$ & $\mathbf{53.78}_{\pm 2.60}$ & $\mathbf{83.52}_{\pm 0.48}$ \\
        \cmidrule(lr){2-9}
        & \multirow{3}{*}{$\epsilon=2$} 
        & \citet{wu2023privacy} & $25.23_{\pm 1.12}$ & $77.66_{\pm 0.19}$ & $40.35_{\pm 0.51}$ & $84.39_{\pm 0.13}$ & $49.59_{\pm 1.20}$ & $83.04_{\pm 0.40}$ \\
        & & \dpsmozo{+On} & $\mathbf{27.66}_{\pm 1.16}$ & $\mathbf{78.24}_{\pm 0.24}$ & $42.56_{\pm 1.14}$ & $84.87_{\pm 0.32}$ & $49.91_{\pm 1.03}$ & $83.28_{\pm 0.36}$ \\
        & & \dpsmozo{+Off} & $26.44_{\pm 1.40}$ & $77.99_{\pm 0.25}$ & $\mathbf{43.62}_{\pm 0.99}$ & $\mathbf{85.03}_{\pm 0.31}$ & $\mathbf{53.45}_{\pm 1.65}$ & $\mathbf{83.40}_{\pm 0.43}$ \\
        \cmidrule(lr){2-9}
        & \multirow{3}{*}{$\epsilon=4$} 
        & \citet{wu2023privacy} & $25.69_{\pm 0.80}$ & $77.67_{\pm 0.30}$ & $40.09_{\pm 0.68}$ & $84.40_{\pm 0.23}$ & $49.58_{\pm 1.13}$ & $82.95_{\pm 0.40}$ \\
        & & \dpsmozo{+On} & $\mathbf{28.20}_{\pm 0.97}$ & $\mathbf{78.34}_{\pm 0.36}$ & $42.98_{\pm 0.22}$ & $\mathbf{85.00}_{\pm 0.24}$ & $49.28_{\pm 0.29}$ & $83.30_{\pm 0.36}$ \\
        & & \dpsmozo{+Off} & $27.47_{\pm 0.96}$ & $77.99_{\pm 0.50}$ & $\mathbf{43.69}_{\pm 1.28}$ & $84.98_{\pm 0.24}$ & $\mathbf{53.93}_{\pm 2.23}$ & $\mathbf{83.44}_{\pm 0.46}$ \\
        \cmidrule(lr){2-9}
        & $\epsilon=\infty$ & Few-shot & $28.62_{\pm 1.42}$ & $78.50_{\pm 0.41}$ & $43.23_{\pm 1.11}$ & $85.08_{\pm 0.27}$ & $49.95_{\pm 1.03}$ & $83.66_{\pm 0.18}$ \\
        \midrule
        
        \multirow{12}{*}{Gemma 2 2B} 
        & $\epsilon=0$ & Zero-shot & $21.77_{\pm 0.52}$ & $74.89_{\pm 0.63}$ & $39.65_{\pm 0.53}$ & $83.66_{\pm 0.27}$ & $41.50_{\pm 0.81}$ & $81.22_{\pm 0.46}$ \\
        \cmidrule(lr){2-9}
        & \multirow{3}{*}{$\epsilon=1$} 
        & \citet{wu2023privacy} & $27.66_{\pm 2.26}$ & $\mathbf{78.25}_{\pm 0.63}$ & $41.92_{\pm 1.39}$ & $85.36_{\pm 0.43}$ & $49.47_{\pm 1.10}$ & $83.04_{\pm 0.40}$ \\
        & & \dpsmozo{+On} & $\mathbf{28.66}_{\pm 0.90}$ & $77.60_{\pm 0.62}$ & $43.20_{\pm 0.56}$ & $\mathbf{85.47}_{\pm 0.51}$ & $50.47_{\pm 0.96}$ & $82.93_{\pm 0.45}$ \\
        & & \dpsmozo{+Off} & $26.13_{\pm 1.36}$ & $76.98_{\pm 0.39}$ & $\mathbf{43.56}_{\pm 1.49}$ & $85.36_{\pm 0.47}$ & $\mathbf{54.32}_{\pm 1.75}$ & $\mathbf{84.08}_{\pm 0.73}$ \\
        \cmidrule(lr){2-9}
        & \multirow{3}{*}{$\epsilon=2$} 
        & \citet{wu2023privacy} & $28.12_{\pm 2.62}$ & $78.25_{\pm 0.62}$ & $41.83_{\pm 0.93}$ & $85.38_{\pm 0.32}$ & $49.65_{\pm 1.30}$ & $83.03_{\pm 0.39}$ \\
        & & \dpsmozo{+On} & $\mathbf{30.09}_{\pm 1.25}$ & $\mathbf{78.79}_{\pm 0.26}$ & $43.47_{\pm 0.95}$ & $\mathbf{85.56}_{\pm 0.46}$ & $50.56_{\pm 0.69}$ & $82.86_{\pm 0.35}$ \\
        & & \dpsmozo{+Off} & $27.10_{\pm 1.54}$ & $77.44_{\pm 0.27}$ & $\mathbf{43.72}_{\pm 1.57}$ & $85.53_{\pm 0.39}$ & $\mathbf{53.51}_{\pm 0.97}$ & $\mathbf{84.28}_{\pm 0.70}$ \\
        \cmidrule(lr){2-9}
        & \multirow{3}{*}{$\epsilon=4$} 
        & \citet{wu2023privacy} & $27.36_{\pm 2.17}$ & $78.28_{\pm 0.71}$ & $42.14_{\pm 1.28}$ & $85.41_{\pm 0.46}$ & $49.83_{\pm 1.33}$ & $83.08_{\pm 0.44}$ \\
        & & \dpsmozo{+On} & $\mathbf{30.43}_{\pm 1.21}$ & $\mathbf{79.21}_{\pm 0.34}$ & $\mathbf{43.71}_{\pm 1.91}$ & $\mathbf{85.69}_{\pm 0.69}$ & $51.14_{\pm 1.24}$ & $83.03_{\pm 0.29}$ \\
        & & \dpsmozo{+Off} & $26.25_{\pm 2.02}$ & $77.36_{\pm 0.81}$ & $43.37_{\pm 0.65}$ & $85.40_{\pm 0.41}$ & $\mathbf{54.00}_{\pm 1.44}$ & $\mathbf{84.27}_{\pm 0.61}$ \\
        \cmidrule(lr){2-9}
        & $\epsilon=\infty$ & Few-shot & $30.39_{\pm 1.34}$ & $79.27_{\pm 0.39}$ & $43.50_{\pm 1.72}$ & $85.51_{\pm 0.73}$ & $52.49_{\pm 0.62}$ & $83.54_{\pm 0.24}$ \\
        \bottomrule
    \end{tabular}}
    \end{center}
\end{table*}

\subsection{Experimental Setup}

\begin{description}[style=unboxed, leftmargin=0cm]
    \item[Datasets.] We evaluate our methodology on three datasets involving diverse generation tasks. The SAMSum dataset \cite{gliwa-etal-2019-samsum} evaluates abstractive summarization, which contains 16k messenger-like conversations with summaries; the E2E dataset \cite{novikova2017e2e} evaluates data-to-text generation, which contains as input some data about a restaurant and generate a sentence in natural language about the restaurant; and the WikiLarge dataset \cite{zhang2017sentence} evaluates text simplification, which consists of complex-simple sentence pairs automatically extracted from English Wikipedia and Simple English Wikipedia by sentence alignment.
    
    \item[Models.] For models, we utilize LLaMA-3 3B \cite{dubey2024llama} and Gemma-2 2B \cite{team2024gemma}, all of which are the instruction fine-tuned version. 
    
    \item[Metrics.] To evaluate the generation quality, we employ the ROUGE-L F-1 \cite{lin2004rouge} and BERTScore-precision \cite{zhang2019bertscore} to measure the lexical and semantic similarity between the LLM-generated and ground-truth output, respectively. We perform each experiment over five seeds, each running on $n_{\text{test}}=100$ test examples to compute the utility (ROUGE-L, BERTScore), then report the mean and standard deviation of the utility across five experiments.  

    For the privacy parameters, we compute the privacy loss of \dpsmozo{} via $(\alpha, \epsilon(\alpha))$-RDP, then convert back to $(\epsilon, \delta)$-DP. We experiment on various $\epsilon$ in the main results, and we follow the standard practice of setting $\delta=1/|D|$. We chose $\alpha$ such that $\epsilon(\alpha) \approx \epsilon/2$, so that the conversion cost from RDP to $(\epsilon, \delta)$-DP is about $\epsilon/2$.

    \item[Baselines.] We compared both \dpsmozo{+On} and \dpsmozo{+Off} with our reimplementation of the Embedding Subspace Aggregation method from \citet{wu2023privacy} using their parameter values at various $\epsilon\in\{1, 2, 4\}$. We also evaluate two additional baselines: $\epsilon=0$, which is the zero-shot outputs, and $\epsilon=\infty$, which is the few-shot ensemble-based decoding without DP. For all methods $n_{\text{shots}}=4$ were used.
    
    More information about the experimental setup (datasets, baselines, hyperparameters) can be found in Appendix \ref{app:add_exp_setup}.
\end{description}

\begin{figure*}[ht!]
    \centering 
    \includegraphics[width=2\columnwidth]{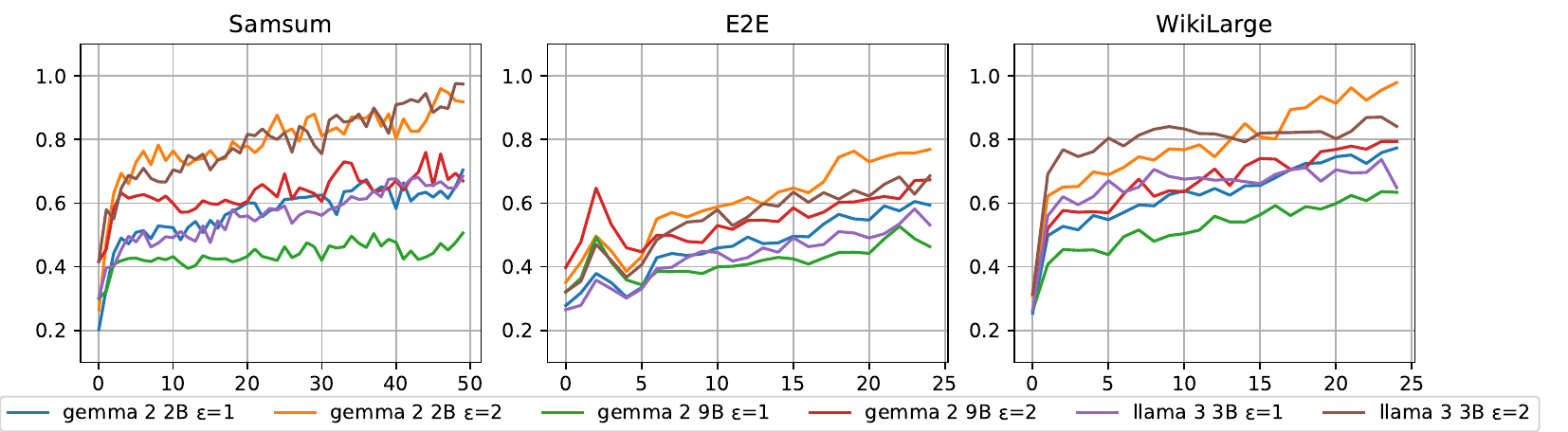}
    \caption{Tracking $\lambda^{(t)}$ from \dpsmozo{+On}, the average value across all test queries of the smallest $\lambda_i$ (y-axis) at the $t$-th generation (x-axis).}
    \label{fig:mean_lambdas}
\end{figure*}

\subsection{Main Results} 
We note that the zero-shot baseline $(\epsilon=0)$ is the ideal privacy where there is no privacy leakage, but it provides a lower bound on the performance since it does not utilize the few-shot examples. In contrast, the non-private few-shot baseline ($\epsilon=\infty)$ is the ideal utility where it fully utilizes the input-output pairs, so it is an upper bound on the performance. However, it provides no rigorous privacy guarantee. For the intermediate $\epsilon$, the performance of \dpsmozo{+On} and \dpsmozo{+Off} should fall somewhere between $\epsilon=0$ and $\epsilon=\infty$. 

We observe this trend in Table \ref{tbl:main_results}, which displays the main results. In particular, \dpsmozo{+On} and \dpsmozo{+Off} substantially improve the performance over zero-shot ICL and match or even exceed the performance of few-shot ICL for strong privacy guarantees. For example, when $\epsilon=2$, \dpsmozo{+On} improves the performance over zero-shot by $8\%$ ROUGE-L F1 and is only $0.3\%$ ROUGE-L less than few-shot performance for SAMSum with Gemma 2 2B. And \dpsmozo{+Off} outperforms few-shot on WikiLarge for both models, demonstrating that LLMs are excellent at generating few-shots which is also observed in \citet{tang2023privacy}

We also observe that for all models and datasets, either \dpsmozo{+On} or \dpsmozo{+Off} outperforms \citet{wu2023privacy}. These results suggest that this baseline overly relies on the zero-shot capabilities of the LLM for generating responses. Hence, their performance is bottlenecked by the zero-shot capabilities of the LLM. Our method, on the other hand, uses a mixture of zero-shot and few-shot outputs for generations to achieve optimal performance. Hence, the performance of our method is bottlenecked by how much the zero-shot and the one-shot outputs diverge from each other, which allows us to achieve strong performance. 

\subsection{Insights Behind the Performance of \dpsmozo{}}
Here, we investigate the inner workings of \dpsmozo{} by measuring the selected mixing parameters $\lambda_i$. Recall that $\lambda_i$ is selected by solving the optimization problem in \cref{eq:lambd_solver}. We are interested in analyzing how the choice of dataset, LLM, and $\epsilon$ affect $\lambda_i$. To this end, we introduce $\lambda_{i, j}^{(t)}$ as the mixing parameter value of the $i$-th example $\lambda_{i}$ evaluated on the $j$-th query at generation iteration $t$. Then define
\begin{equation*}
    \textstyle \lambda^{(t)} = \frac{1}{|D_{\text{test}}|}\sum_{j=1}^{|D_{\text{test}}|} \min_{0\leq i \leq n_{\text{shots}}} \lambda_{i, j}^{(t)}.
\end{equation*}
Hence, $\lambda^{(t)}$ reports the average value across all test queries of the smallest $\lambda_i$ at the $t$-th generation. 

\cref{fig:mean_lambdas} plots $\lambda^{(t)}$ from \dpsmozo{+On} evaluated on SAMSum, E2E, and Spider. A clear trend emerges for all models, datasets, and $\epsilon$: the $\lambda^{(t)}$ is lowest during the first generation, spikes around the third or fourth token generation, then slowly increases afterward. We attribute this behavior to the fact that, in the beginning, the LLM has no prior generations, so it overly relies on its prior knowledge and/or the input-output examples. Hence why the one-shot and zero-shot diverge so much, causing the resulting $\lambda^{(t)}$ to be small. However, once the LLM generates more tokens, it relies on its previous output tokens more to generate the next token, regardless of its prior knowledge and the input-output mapping. Hence why the one-shot and zero-shot converge. 

Another interesting finding we observed is for a given $\epsilon$, generally $\lambda^{(t)}$ is larger for Gemma 2 2B than for LLama 3 3B. This indicates that the demonstrations do not cause the one-shot outputs to diverge too much from the zero-shot output for Gemma 2 2B. Hence, Gemma 2 2B can rely on its parametric knowledge more than LLama 3 3B. Additionally, $\lambda^{(t)}$ is generally smaller for Gemma 2 9B than Gemma 2 2B, so larger models are influenced by ICL demonstrations more than smaller ones. 

\subsection{Ablation Studies}
\begin{figure*}[ht!]
    \centering 
    \includegraphics[width=2\columnwidth]{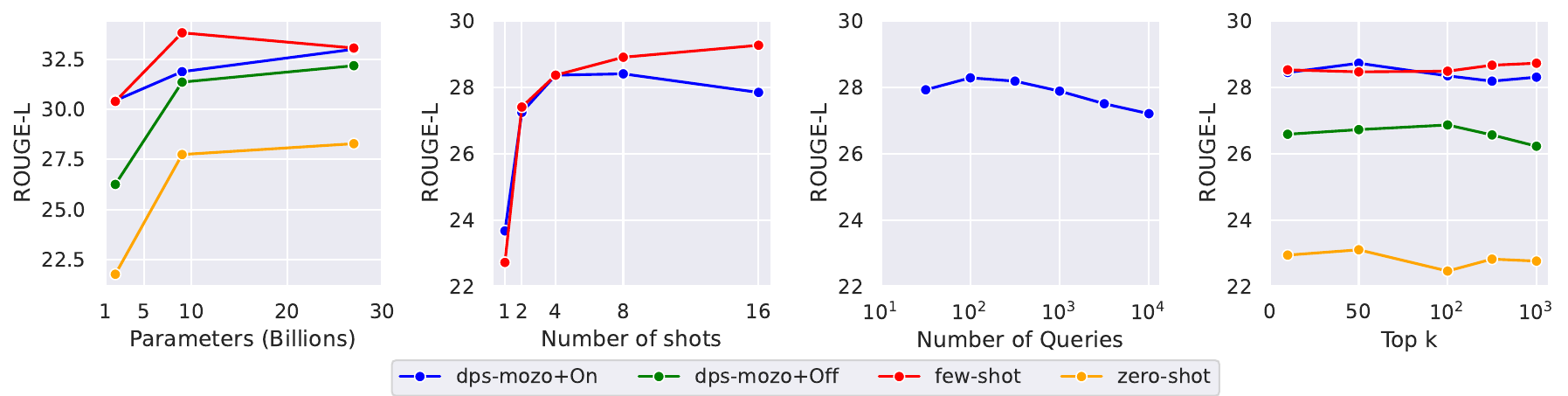}
    \caption{Ablation studies on dialogue summarization using the SAMSum dataset.}
    \label{fig:ablations}
\end{figure*}

Next, we conduct an ablation study on dialogue summarization using the Samsum dataset. Unless stated otherwise, each ablation uses LLaMA 3 3B, $\epsilon=4$, $n_{\text{test}}=100$, $k=100$, and $n_{\text{shots}}=4$. \cref{fig:ablations} shows the results for model size (parameters) $|\theta|$, number of shots $n_{\text{shots}}$, number of queries $n_{\text{test}}$, and top-$k$.

\textbf{Model Size.} We evaluate the performance of \dpsmozo{+On} and \dpsmozo{+Off} on various model sizes of Gemma 2, which are 2B, 9B, and 27B. We see that increasing the model size improves the performance of both methods. Impressively, \dpsmozo{+Off} benefits substantially from the increased model size as its performance approaches \dpsmozo{+On} for 9B and 27B.

\textbf{Number Shots.} Next, we evaluate the performance of \dpsmozo{+On} on various number of shots. We observe that for $n_{\text{shots}} \leq 4$, \dpsmozo{+On} perform as well as few-shot. However, for a larger number of shots, we see that the performance of \dpsmozo{+On} degrades. We surmise that there are marginal performance gains when $n_{\text{shots}} > 4$ for few-shot ICL, and there is a decrease in privacy amplification by subsampling for larger $n_{\text{shots}}$. Hence, both factors reduce the performance of \dpsmozo{+On}.

\textbf{Queries.} Here, we evaluate the performance of \dpsmozo{+On} when scaling the number of queries. Our results demonstrate that even when scaling the number of queries by two orders of magnitude, the performance of \dpsmozo{+On} degrades marginally; more precisely, roughly $1\%$ ROUGE-L decrease going from $10^2$ to $10^4$.

\textbf{Top-k}. Lastly, we evaluate the impact of top-k truncation on \dpsmozo{+On} and \dpsmozo{+Off}. The results show that the performance of the few-shot improves with larger $k\geq 50$ while the performance of \dpsmozo{+On} and \dpsmozo{+Off} degrades. We attribute this to the fact that truncating the output space helps converge the one-shot and zero-shot outputs, leading to higher $\lambda_i$.

\subsection{Empirical Privacy Evaluation}
\begin{table}[t!]
    \caption{AUC-ROC scores for membership inference attacks under different privacy budgets ($\epsilon$) for AG News and Trec with LLaMA-3 3B. $\epsilon=\infty$ represents standard few-shot learning without privacy protection.}
    \label{tbl:MIA_results}
    \begin{center}
    \resizebox{\linewidth}{!}{%
    \begin{tabular}{l|c|c|c|c|c}
        \toprule 
        Dataset & $\epsilon=1$ & $\epsilon=2$ & $\epsilon=4$ & $\epsilon=8$ & $\epsilon=\infty$ \\
        \midrule
        AG News & 50.41 & 50.65 & 51.08 & 51.84 & 78.20 \\
        \midrule
        Trec & 50.30 & 50.45 & 50.72 & 51.16 & 75.72 \\
        \bottomrule
    \end{tabular}}
    \end{center}
\end{table}

We evaluate the empirical privacy leakage of \dpsmozo{+On} through Membership Inference Attacks (MIAs) \cite{shokri2017membership}. We use the MIA setup from \citet{duan2024flocks}, which used text classification tasks on AG News \cite{zhang2015character} and Trec \cite{voorhees2000building}. Following \citet{yeom2018privacy}, we define the membership inference score as the probability of the model outputting the correct label. In our setting, this score is computed as $p_\theta(y_{\text{true}}|\mathbf{x}', \mathbf{y}', \mathbf{x}_i)$, where $\mathbf{x}_i\in D_{\text{test}}$ is the query example, $y_{\text{true}}$ is the ground-truth label, and $(\mathbf{x}', \mathbf{y}')$ is the member example. Please refer to more details in \cref{app:mia_setup}.

\cref{tbl:MIA_results} displays the AUC-ROC score for \dpsmozo{+On} with LLaMA-3 3B on AG News and Trec. A score of 50\% for AUC-ROC is ideal because an adversary can only perform as well as random guessing. We see that for small $\epsilon$, the adversary struggles to obtain useful information, indicating that our framework can protect against this type of privacy attack. However, without privacy guarantee $\epsilon=\infty$, an adversary can obtain strong information leakage, motivating the need for privacy safeguards in the ICL domain. 

\section{Discussion and Conclusion}
In this work, we proposed \dpsmozo{}, a differentially private decoding framework that mixes one-shots with zero-shot output distributions and then samples from it. By utilizing the inherent stochasticity of language models and adaptively controlling the amount of privacy that is leaked by ICL demonstrations, we can avoid additive noise commonly found in DP solutions and achieve strong privacy-utility tradeoffs, as demonstrated by our experimental evaluations. We conclude with crucial implications about \dpsmozo{}.

Since \dpsmozo{} needs access to the output probability of the LLM, we could not experiment with closed-source LLMs, such as GPT-4, due to APIs returning log probabilities for only the top $k$ tokens. It is possible to uncover the full logit vector by utilizing the logit bias \cite{carlini2024stealing}, which we could then apply $\dpsmozo{}$. We leave this as future work. Moreover, our problem setup and \dpsmozo{} are intended for LLM providers, so it is fair to assume access to the entire output probability of the model. 

However, \dpsmozo{} assumes that an adversary only has access to the sampled output tokens. This is one of the reasons why \dpsmozo{} has strong privacy-utility tradeoffs since it does not need to privatize the entire next-token output distribution, which can be upwards of 50,000 dimensions. This provides better utility while satisfying a practical threat model, as APIs already limit the returned output logits. Consequently, the LLM provider cannot fully release the mixed output distribution from \cref{eq:mix}.

\section*{Impact Statement}
With the rapid adoption of large language models (LLMs) and the exponential increase in their use for in-context learning (ICL), ensuring user privacy has become a critical priority. This work offers robust and rigorous privacy guarantees, paving the way for private utilization of LLMs in ICL queries. By safeguarding user data with differential privacy, this research enables the responsible and privacy-conscious deployment of LLMs, fostering trust and expanding their potential for transformative applications.


\newpage
\bibliography{main}

\begin{thebibliography}{49}
\providecommand{\natexlab}[1]{#1}
\providecommand{\url}[1]{\texttt{#1}}
\expandafter\ifx\csname urlstyle\endcsname\relax
  \providecommand{\doi}[1]{doi: #1}\else
  \providecommand{\doi}{doi: \begingroup \urlstyle{rm}\Url}\fi

\bibitem[Balle et~al.(2018)Balle, Barthe, and Gaboardi]{balle2018privacy}
Balle, B., Barthe, G., and Gaboardi, M.
\newblock Privacy amplification by subsampling: Tight analyses via couplings and divergences.
\newblock \emph{Advances in neural information processing systems}, 31, 2018.

\bibitem[Balle et~al.(2020)Balle, Barthe, Gaboardi, Hsu, and Sato]{balle2020hypothesis}
Balle, B., Barthe, G., Gaboardi, M., Hsu, J., and Sato, T.
\newblock Hypothesis testing interpretations and renyi differential privacy.
\newblock In \emph{International Conference on Artificial Intelligence and Statistics}, pp.\  2496--2506. PMLR, 2020.

\bibitem[Brown et~al.(2020)Brown, Mann, Ryder, Subbiah, Kaplan, Dhariwal, Neelakantan, Shyam, Sastry, Askell, et~al.]{brown2020language}
Brown, T., Mann, B., Ryder, N., Subbiah, M., Kaplan, J.~D., Dhariwal, P., Neelakantan, A., Shyam, P., Sastry, G., Askell, A., et~al.
\newblock Language models are few-shot learners.
\newblock \emph{Advances in neural information processing systems}, 33:\penalty0 1877--1901, 2020.

\bibitem[Carlini et~al.(2024)Carlini, Paleka, Dvijotham, Steinke, Hayase, Cooper, Lee, Jagielski, Nasr, Conmy, et~al.]{carlini2024stealing}
Carlini, N., Paleka, D., Dvijotham, K.~D., Steinke, T., Hayase, J., Cooper, A.~F., Lee, K., Jagielski, M., Nasr, M., Conmy, A., et~al.
\newblock Stealing part of a production language model.
\newblock \emph{arXiv preprint arXiv:2403.06634}, 2024.

\bibitem[Devlin et~al.(2018)Devlin, Chang, Lee, and Toutanova]{devlin2018bert}
Devlin, J., Chang, M.-W., Lee, K., and Toutanova, K.
\newblock Bert: Pre-training of deep bidirectional transformers for language understanding.
\newblock \emph{arXiv preprint arXiv:1810.04805}, 2018.

\bibitem[Duan et~al.(2023)Duan, Dziedzic, Yaghini, Papernot, and Boenisch]{duan2023privacy}
Duan, H., Dziedzic, A., Yaghini, M., Papernot, N., and Boenisch, F.
\newblock On the privacy risk of in-context learning.
\newblock In \emph{The 61st Annual Meeting Of The Association For Computational Linguistics}, 2023.

\bibitem[Duan et~al.(2024)Duan, Dziedzic, Papernot, and Boenisch]{duan2024flocks}
Duan, H., Dziedzic, A., Papernot, N., and Boenisch, F.
\newblock Flocks of stochastic parrots: Differentially private prompt learning for large language models.
\newblock \emph{Advances in Neural Information Processing Systems}, 36, 2024.

\bibitem[Dubey et~al.(2024)Dubey, Jauhri, Pandey, Kadian, Al-Dahle, Letman, Mathur, Schelten, Yang, Fan, et~al.]{dubey2024llama}
Dubey, A., Jauhri, A., Pandey, A., Kadian, A., Al-Dahle, A., Letman, A., Mathur, A., Schelten, A., Yang, A., Fan, A., et~al.
\newblock The llama 3 herd of models.
\newblock \emph{arXiv preprint arXiv:2407.21783}, 2024.

\bibitem[Dwork(2006)]{dwork2006differential}
Dwork, C.
\newblock Differential privacy.
\newblock In \emph{International colloquium on automata, languages, and programming}, pp.\  1--12. Springer, 2006.

\bibitem[Dwork \& Feldman(2018)Dwork and Feldman]{dwork2018privacy}
Dwork, C. and Feldman, V.
\newblock Privacy-preserving prediction.
\newblock In \emph{Conference On Learning Theory}, pp.\  1693--1702. PMLR, 2018.

\bibitem[Dwork et~al.(2014)Dwork, Roth, et~al.]{dwork2014algorithmic}
Dwork, C., Roth, A., et~al.
\newblock The algorithmic foundations of differential privacy.
\newblock \emph{Foundations and Trends{\textregistered} in Theoretical Computer Science}, 9\penalty0 (3--4):\penalty0 211--407, 2014.

\bibitem[Fan et~al.(2018)Fan, Lewis, and Dauphin]{fan2018hierarchical}
Fan, A., Lewis, M., and Dauphin, Y.
\newblock Hierarchical neural story generation.
\newblock \emph{arXiv preprint arXiv:1805.04833}, 2018.

\bibitem[Flemings et~al.(2024)Flemings, Razaviyayn, and Annavaram]{flemings2024differentially}
Flemings, J., Razaviyayn, M., and Annavaram, M.
\newblock Differentially private next-token prediction of large language models.
\newblock \emph{arXiv preprint arXiv:2403.15638}, 2024.

\bibitem[Ginart et~al.(2022)Ginart, van~der Maaten, Zou, and Guo]{ginart2022submix}
Ginart, A., van~der Maaten, L., Zou, J., and Guo, C.
\newblock Submix: Practical private prediction for large-scale language models.
\newblock \emph{arXiv preprint arXiv:2201.00971}, 2022.

\bibitem[Gliwa et~al.(2019)Gliwa, Mochol, Biesek, and Wawer]{gliwa-etal-2019-samsum}
Gliwa, B., Mochol, I., Biesek, M., and Wawer, A.
\newblock {SAMS}um corpus: A human-annotated dialogue dataset for abstractive summarization.
\newblock In \emph{Proceedings of the 2nd Workshop on New Frontiers in Summarization}, pp.\  70--79, Hong Kong, China, November 2019. Association for Computational Linguistics.
\newblock \doi{10.18653/v1/D19-5409}.
\newblock URL \url{https://www.aclweb.org/anthology/D19-5409}.

\bibitem[Hanke et~al.(2024)Hanke, Blanchard, Boenisch, Olatunji, Backes, and Dziedzic]{hanke2024open}
Hanke, V., Blanchard, T., Boenisch, F., Olatunji, I.~E., Backes, M., and Dziedzic, A.
\newblock Open llms are necessary for current private adaptations and outperform their closed alternatives.
\newblock \emph{arXiv preprint arXiv:2411.05818}, 2024.

\bibitem[Hong et~al.(2023)Hong, Wang, Zhang, Li, Li, and Wang]{hong2023dp}
Hong, J., Wang, J.~T., Zhang, C., Li, Z., Li, B., and Wang, Z.
\newblock Dp-opt: Make large language model your privacy-preserving prompt engineer.
\newblock \emph{arXiv preprint arXiv:2312.03724}, 2023.

\bibitem[Husain et~al.(2020)Husain, Balle, Cranko, and Nock]{husain2020local}
Husain, H., Balle, B., Cranko, Z., and Nock, R.
\newblock Local differential privacy for sampling.
\newblock In \emph{International Conference on Artificial Intelligence and Statistics}, pp.\  3404--3413. PMLR, 2020.

\bibitem[Li \& Li(2023)Li and Li]{li2023angle}
Li, X. and Li, J.
\newblock Angle-optimized text embeddings.
\newblock \emph{arXiv preprint arXiv:2309.12871}, 2023.

\bibitem[Lin(2004)]{lin2004rouge}
Lin, C.-Y.
\newblock Rouge: A package for automatic evaluation of summaries.
\newblock In \emph{Text summarization branches out}, pp.\  74--81, 2004.

\bibitem[Majmudar et~al.(2022)Majmudar, Dupuy, Peris, Smaili, Gupta, and Zemel]{majmudar2022differentially}
Majmudar, J., Dupuy, C., Peris, C., Smaili, S., Gupta, R., and Zemel, R.
\newblock Differentially private decoding in large language models.
\newblock \emph{arXiv preprint arXiv:2205.13621}, 2022.

\bibitem[Min et~al.(2021)Min, Lewis, Hajishirzi, and Zettlemoyer]{min2021noisy}
Min, S., Lewis, M., Hajishirzi, H., and Zettlemoyer, L.
\newblock Noisy channel language model prompting for few-shot text classification.
\newblock \emph{arXiv preprint arXiv:2108.04106}, 2021.

\bibitem[Min et~al.(2022)Min, Lyu, Holtzman, Artetxe, Lewis, Hajishirzi, and Zettlemoyer]{min2022rethinking}
Min, S., Lyu, X., Holtzman, A., Artetxe, M., Lewis, M., Hajishirzi, H., and Zettlemoyer, L.
\newblock Rethinking the role of demonstrations: What makes in-context learning work?
\newblock \emph{arXiv preprint arXiv:2202.12837}, 2022.

\bibitem[Mironov(2017)]{mironov2017renyi}
Mironov, I.
\newblock R{\'e}nyi differential privacy.
\newblock In \emph{2017 IEEE 30th computer security foundations symposium (CSF)}, pp.\  263--275. IEEE, 2017.

\bibitem[Nissim et~al.(2007)Nissim, Raskhodnikova, and Smith]{nissim2007smooth}
Nissim, K., Raskhodnikova, S., and Smith, A.
\newblock Smooth sensitivity and sampling in private data analysis.
\newblock In \emph{Proceedings of the thirty-ninth annual ACM symposium on Theory of computing}, pp.\  75--84, 2007.

\bibitem[Novikova et~al.(2017)Novikova, Du{\v{s}}ek, and Rieser]{novikova2017e2e}
Novikova, J., Du{\v{s}}ek, O., and Rieser, V.
\newblock The e2e dataset: New challenges for end-to-end generation.
\newblock \emph{arXiv preprint arXiv:1706.09254}, 2017.

\bibitem[Papernot et~al.(2016)Papernot, Abadi, Erlingsson, Goodfellow, and Talwar]{papernot2016semi}
Papernot, N., Abadi, M., Erlingsson, U., Goodfellow, I., and Talwar, K.
\newblock Semi-supervised knowledge transfer for deep learning from private training data.
\newblock \emph{arXiv preprint arXiv:1610.05755}, 2016.

\bibitem[Papernot et~al.(2018)Papernot, Song, Mironov, Raghunathan, Talwar, and Erlingsson]{papernot2018scalable}
Papernot, N., Song, S., Mironov, I., Raghunathan, A., Talwar, K., and Erlingsson, {\'U}.
\newblock Scalable private learning with pate.
\newblock \emph{arXiv preprint arXiv:1802.08908}, 2018.

\bibitem[Priyanshu et~al.(2023)Priyanshu, Vijay, Kumar, Naidu, and Mireshghallah]{priyanshu2023chatbots}
Priyanshu, A., Vijay, S., Kumar, A., Naidu, R., and Mireshghallah, F.
\newblock Are chatbots ready for privacy-sensitive applications? an investigation into input regurgitation and prompt-induced sanitization.
\newblock \emph{arXiv preprint arXiv:2305.15008}, 2023.

\bibitem[Radford et~al.(2019)Radford, Wu, Child, Luan, Amodei, Sutskever, et~al.]{radford2019language}
Radford, A., Wu, J., Child, R., Luan, D., Amodei, D., Sutskever, I., et~al.
\newblock Language models are unsupervised multitask learners.
\newblock \emph{OpenAI blog}, 1\penalty0 (8):\penalty0 9, 2019.

\bibitem[Shokri et~al.(2017)Shokri, Stronati, Song, and Shmatikov]{shokri2017membership}
Shokri, R., Stronati, M., Song, C., and Shmatikov, V.
\newblock Membership inference attacks against machine learning models.
\newblock In \emph{2017 IEEE symposium on security and privacy (SP)}, pp.\  3--18. IEEE, 2017.

\bibitem[Steinke(2022)]{steinke2022composition}
Steinke, T.
\newblock Composition of differential privacy \& privacy amplification by subsampling.
\newblock \emph{arXiv preprint arXiv:2210.00597}, 2022.

\bibitem[Tang et~al.(2023)Tang, Shin, Inan, Manoel, Mireshghallah, Lin, Gopi, Kulkarni, and Sim]{tang2023privacy}
Tang, X., Shin, R., Inan, H.~A., Manoel, A., Mireshghallah, F., Lin, Z., Gopi, S., Kulkarni, J., and Sim, R.
\newblock Privacy-preserving in-context learning with differentially private few-shot generation.
\newblock \emph{arXiv preprint arXiv:2309.11765}, 2023.

\bibitem[Team et~al.(2024)Team, Riviere, Pathak, Sessa, Hardin, Bhupatiraju, Hussenot, Mesnard, Shahriari, Ram{\'e}, et~al.]{team2024gemma}
Team, G., Riviere, M., Pathak, S., Sessa, P.~G., Hardin, C., Bhupatiraju, S., Hussenot, L., Mesnard, T., Shahriari, B., Ram{\'e}, A., et~al.
\newblock Gemma 2: Improving open language models at a practical size.
\newblock \emph{arXiv preprint arXiv:2408.00118}, 2024.

\bibitem[Tian et~al.(2022)Tian, Zhao, Huang, Wang, Zhang, and He]{tian2022seqpate}
Tian, Z., Zhao, Y., Huang, Z., Wang, Y.-X., Zhang, N.~L., and He, H.
\newblock Seqpate: Differentially private text generation via knowledge distillation.
\newblock \emph{Advances in Neural Information Processing Systems}, 35:\penalty0 11117--11130, 2022.

\bibitem[Van~Erven \& Harremos(2014)Van~Erven and Harremos]{van2014renyi}
Van~Erven, T. and Harremos, P.
\newblock R{\'e}nyi divergence and kullback-leibler divergence.
\newblock \emph{IEEE Transactions on Information Theory}, 60\penalty0 (7):\penalty0 3797--3820, 2014.

\bibitem[Voorhees \& Tice(2000)Voorhees and Tice]{voorhees2000building}
Voorhees, E.~M. and Tice, D.~M.
\newblock Building a question answering test collection.
\newblock In \emph{Proceedings of the 23rd annual international ACM SIGIR conference on Research and development in information retrieval}, pp.\  200--207, 2000.

\bibitem[Wang et~al.(2023)Wang, Chen, Pei, Xie, Kang, Zhang, Xu, Xiong, Dutta, Schaeffer, et~al.]{wang2023decodingtrust}
Wang, B., Chen, W., Pei, H., Xie, C., Kang, M., Zhang, C., Xu, C., Xiong, Z., Dutta, R., Schaeffer, R., et~al.
\newblock Decodingtrust: A comprehensive assessment of trustworthiness in gpt models.
\newblock In \emph{NeurIPS}, 2023.

\bibitem[Wang et~al.(2015)Wang, Fienberg, and Smola]{wang2015privacy}
Wang, Y.-X., Fienberg, S., and Smola, A.
\newblock Privacy for free: Posterior sampling and stochastic gradient monte carlo.
\newblock In \emph{International Conference on Machine Learning}, pp.\  2493--2502. PMLR, 2015.

\bibitem[Wang et~al.(2019)Wang, Balle, and Kasiviswanathan]{wang2019subsampled}
Wang, Y.-X., Balle, B., and Kasiviswanathan, S.~P.
\newblock Subsampled r{\'e}nyi differential privacy and analytical moments accountant.
\newblock In \emph{The 22nd international conference on artificial intelligence and statistics}, pp.\  1226--1235. PMLR, 2019.

\bibitem[Wei et~al.(2023)Wei, Wei, Tay, Tran, Webson, Lu, Chen, Liu, Huang, Zhou, et~al.]{wei2023larger}
Wei, J., Wei, J., Tay, Y., Tran, D., Webson, A., Lu, Y., Chen, X., Liu, H., Huang, D., Zhou, D., et~al.
\newblock Larger language models do in-context learning differently.
\newblock \emph{arXiv preprint arXiv:2303.03846}, 2023.

\bibitem[Wu et~al.(2023)Wu, Panda, Wang, and Mittal]{wu2023privacy}
Wu, T., Panda, A., Wang, J.~T., and Mittal, P.
\newblock Privacy-preserving in-context learning for large language models.
\newblock \emph{arXiv preprint arXiv:2305.01639}, 2023.

\bibitem[Xie et~al.(2024)Xie, Lin, Backurs, Gopi, Yu, Inan, Nori, Jiang, Zhang, Lee, et~al.]{xie2024differentially}
Xie, C., Lin, Z., Backurs, A., Gopi, S., Yu, D., Inan, H.~A., Nori, H., Jiang, H., Zhang, H., Lee, Y.~T., et~al.
\newblock Differentially private synthetic data via foundation model apis 2: Text.
\newblock \emph{arXiv preprint arXiv:2403.01749}, 2024.

\bibitem[Xu(2023)]{xu2023context}
Xu, Z.
\newblock Context-aware decoding reduces hallucination in query-focused summarization.
\newblock \emph{arXiv preprint arXiv:2312.14335}, 2023.

\bibitem[Yeom et~al.(2018)Yeom, Giacomelli, Fredrikson, and Jha]{yeom2018privacy}
Yeom, S., Giacomelli, I., Fredrikson, M., and Jha, S.
\newblock Privacy risk in machine learning: Analyzing the connection to overfitting.
\newblock In \emph{2018 IEEE 31st computer security foundations symposium (CSF)}, pp.\  268--282. IEEE, 2018.

\bibitem[Zhang et~al.(2019)Zhang, Kishore, Wu, Weinberger, and Artzi]{zhang2019bertscore}
Zhang, T., Kishore, V., Wu, F., Weinberger, K.~Q., and Artzi, Y.
\newblock Bertscore: Evaluating text generation with bert.
\newblock \emph{arXiv preprint arXiv:1904.09675}, 2019.

\bibitem[Zhang \& Lapata(2017)Zhang and Lapata]{zhang2017sentence}
Zhang, X. and Lapata, M.
\newblock Sentence simplification with deep reinforcement learning.
\newblock \emph{arXiv preprint arXiv:1703.10931}, 2017.

\bibitem[Zhang et~al.(2015)Zhang, Zhao, and LeCun]{zhang2015character}
Zhang, X., Zhao, J., and LeCun, Y.
\newblock Character-level convolutional networks for text classification.
\newblock \emph{Advances in neural information processing systems}, 28, 2015.

\bibitem[Zhao et~al.(2021)Zhao, Wallace, Feng, Klein, and Singh]{zhao2021calibrate}
Zhao, Z., Wallace, E., Feng, S., Klein, D., and Singh, S.
\newblock Calibrate before use: Improving few-shot performance of language models.
\newblock In \emph{International conference on machine learning}, pp.\  12697--12706. PMLR, 2021.

\end{thebibliography}
\bibliographystyle{icml2025}

\newpage
\appendix
\onecolumn
\section{Properties of RDP}\label{sec:useful_thms}
In this section, we provide some properties on RDP that will be relevant to proving that \dpsmozo{} satisfies RDP.

\begin{theorem}[Post-Processing \cite{mironov2017renyi}]\label{thm:post-processing}
   Let $A: \mathcal{D} \rightarrow \mathcal{R}$ be $(\alpha, \epsilon(\alpha))$-RDP, and let $F: \mathcal{R} \rightarrow \mathcal{Z}$ be an arbitrary randomized mapping. Then $F \circ M$ is $(\alpha, \epsilon(\alpha))$-RDP.
\end{theorem}

\begin{theorem}[Composition \cite{mironov2017renyi}]\label{thm:composition}
    Let $A_1, ..., A_k$ be a sequence of $(\alpha, \epsilon(\alpha))$-RDP algorithms. Then the composition $A_k \circ A_{k-1} \circ ... \circ A_1$ is $(\alpha, k\epsilon(\alpha))$-RDP.
\end{theorem}

\begin{theorem}[Conversion from RDP to Approximate DP \cite{balle2020hypothesis}]\label{thm:rdp-dp}
    If an algorithm $A$ is $(\alpha, \epsilon(\alpha))$-RDP, then it is $(\epsilon(\alpha) + \log((\alpha-1)/ \alpha) - (\log\delta + \log \alpha)/(\alpha-1), \delta)$-DP for any $0 < \delta < 1$.
\end{theorem}

\begin{theorem}[RDP for Subsampled Mechanisms \cite{wang2019subsampled}.]\label{thm:amp_subsamp}
   Given a dataset of $n$ points drawn from a domain $\mathcal{D}$ and an algorithm $\mathcal{A}$ that takes an input from $\mathcal{D}^m$ for $m \leq n$, let the randomized algorithm $\mathcal{A} \circ \textbf{subsample}$ be defined as: (1) subsample without replacement $m$ datapoints of the dataset (sampling parameter $q=m/n$), and (2) apply $\mathcal{A}$ to the sbusampled dataset. For all integers $\alpha \geq 2$, if $\mathcal{A}$ obeys $(\alpha, \epsilon(\alpha))$-RDP, then the subsampled mechanism $\mathcal{A}\circ\textbf{subsample}$ obeys $(\alpha, \epsilon'(\alpha))$-RDP where, 
   \begin{dmath*}
       \epsilon'(\alpha) \leq \frac{1}{\alpha-1} \log \left(1 + q^2 {\alpha\choose 2}\min \left\{4(e^{\epsilon(2)}-1), e^{\epsilon(2)}\min\left\{2, (e^{\epsilon(\infty)} - 1)^2 \right\}\right\} \\
       + \sum_{j=3}^{\alpha}q^j {\alpha\choose j}e^{(j-1)\epsilon(j)}\min\left\{2, (e^{\epsilon(\infty)}-1)^j \right\}  \right).
   \end{dmath*}
\end{theorem}

\section{Detailed Privacy Analysis of \dpsmozo{}}
\label{sec:priv_anlaysis}
We now provide a rigorous privacy analysis of \dpsmozo{}. First, we introduce a couple of properties of Renyi Divergence to help with the proof.

\begin{theorem}[Triangle-like inequality, lemma 33.7 from \citet{steinke2022composition}]\label{thm:weak-triangle-inequality}
   Let $P, Q, R$ be distributions on $\mathcal{R}$. If $\renyidiv{\alpha}{P}{Q} \leq \epsilon_1 \alpha$ and $\renyidiv{\alpha}{Q}{R} \leq \epsilon_2 \alpha$ for $1 < \alpha < \infty$, then 
   \begin{equation}
        \renyidiv{\alpha}{P}{R} \leq (\sqrt{\epsilon_1} + \sqrt{\epsilon_2})^{2} \alpha.
   \end{equation}
\end{theorem}

\begin{theorem}[Additivity, Theorem 28 from \citet{van2014renyi}]\label{lemma:renyi_div_indep}
For $n=1, 2, ...$, let $(P_n, Q_n)$ be pairs of probability distributions on measurable spaces $(\mathcal{X}_n, \mathcal{F}_n)$. Then for any $\alpha\in [0, \infty]$ and any $N\in\{1, 2, ...\}$ 
\begin{equation}
    \sum_{n=1}^{N} \renyidiv{\alpha}{P_n}{Q_n} = \renyidiv{\alpha}{P_1 \times ... \times P_N}{Q_1 \times ... \times Q_N}
\end{equation}
\end{theorem}

Now we show that \dpsmozo{} achieves DP for the generated next token:

\begin{theorem}
     Let $\mathcal{A}$ to denote \dpsmozo{} (\cref{alg:dpsmozo}). Let $D=\left\{(\mathbf{x}_j, \mathbf{y}_j)\right\}_{j=1}^{n}$ be a dataset comprised of input-output examples, $\mathbf{x}$ be an input query, and $\mathbf{y}_{<t}$ be previous $t-1$ generated DP tokens. Then the next token $y_t \sim \mathcal{A}(D, \mathbf{x}, \mathbf{y}_{<t})$ generated by $\mathcal{A}$ is $(\alpha, 4\beta\alpha)$-RDP. 
\end{theorem}

\textit{Proof.} Consider two adjacent datasets $D, D'$ such that they differ by only one instance. Without loss of generality, suppose it is the $i$-th instance. Note that since $\mathbf{y}_{<t}$ is DP, then $p_{\theta}(\mathbf{x}, \mathbf{y}_{<t})$ does not leak additional information about $D$ by the post-processing of DLP (\cref{thm:post-processing}). Working through the Renyi Divergence between the output of Algorithm 1 with $D$ and $D'$ gives the following privacy loss for generating the next token.

\begin{align}
    &\renyidiv{\alpha}*{\mathcal{A}(D, \mathbf{x}, \mathbf{y}_{<t})}{\mathcal{A}(D', \mathbf{x}, \mathbf{y}_{<t})} \nonumber \\
    &=\renyidiv{\alpha}*{\prod_{j\neq i}\overline{p}_{\theta, \lambda_j}((\mathbf{x}_j, \mathbf{y}_j), \mathbf{x}, \mathbf{y}_{<t}) \overline{p}_{\theta, \lambda_i}((\mathbf{x}_i, \mathbf{y}_i), \mathbf{x}, \mathbf{y}_{<t})}{\prod_{j\neq i}\overline{p}_{\theta, \lambda_j}((\mathbf{x}_j, \mathbf{y}_j), \mathbf{x}, \mathbf{y}_{<t}) \overline{p}_{\theta, \lambda_i}((\mathbf{x}'_i, \mathbf{y}'_i), \mathbf{x}, \mathbf{y}_{<t})} \nonumber \\
    &=\renyidiv{\alpha}*{\prod_{j\neq i}\overline{p}_{\theta, \lambda_j}((\mathbf{x}_j, \mathbf{y}_j), \mathbf{x}, \mathbf{y}_{<t})}{\prod_{j\neq i}\overline{p}_{\theta, \lambda_j}((\mathbf{x}_j, \mathbf{y}_j), \mathbf{x}, \mathbf{y}_{<t})} \nonumber \\
    &+ \renyidiv{\alpha}*{\overline{p}_{\theta, \lambda_i}((\mathbf{x}_i, \mathbf{y}_i), \mathbf{x}, \mathbf{y}_{<t})}{\overline{p}_{\theta, \lambda_i}((\mathbf{x}'_i, \mathbf{y}'_i), \mathbf{x}, \mathbf{y}_{<t})} \label{eq:use_lemma} \\
    &= 0 + \renyidiv{\alpha}{\overline{p}_{\theta, \lambda_i}((\mathbf{x}_i, \mathbf{y}_i), \mathbf{x}, \mathbf{y}_{<t})}{\overline{p}_{\theta, \lambda_i}((\mathbf{x}'_i, \mathbf{y}'_i), \mathbf{x}, \mathbf{y}_{<t})} \nonumber \\
    &\leq (\sqrt{\beta} + \sqrt{\beta})^2\alpha \label{eq:use_tri_ineq} \\
    &= 4\beta\alpha \nonumber
\end{align}

where Eq. \ref{eq:use_lemma} uses Lemma \ref{lemma:renyi_div_indep} and Eq. \ref{eq:use_tri_ineq} uses the Triangle-like Inequality (Theorem \ref{thm:weak-triangle-inequality}) since $\renyidiv{\alpha}{\overline{p}_{\theta, \lambda_i}((\mathbf{x}_i, \mathbf{y}_i), \mathbf{x}, \mathbf{y}_{<t})}{p_{\theta}(\mathbf{x}, \mathbf{y}_{<t})} \leq \beta\alpha$ and $\renyidiv{\alpha}{p_{\theta}(\mathbf{x}, \mathbf{y}_{<t})}{\overline{p}_{\theta, \lambda_j}((\mathbf{x}'_i, \mathbf{y}'_i), \mathbf{x}, \mathbf{y}_{<t})}\leq\beta\alpha$ from line \ref{line:set_lambd} of Algorithm \ref{alg:dpsmozo}. \qed

Now that we have shown that Algorithm \ref{alg:dpsmozo} is RDP, we can perform subsampling on the private dataset to obtain a subset $S_t\subseteq D$ to use for the algorithm. Hence, we can invoke the privacy amplification by subsampling (Theorem \ref{thm:amp_subsamp}) to further strengthen our privacy analysis. The resulting privacy loss $\epsilon'(\alpha)$ will be substantially smaller than $\epsilon(\alpha)=4\beta\alpha$, so we select the $\beta$ for a fixed $\alpha$ by solving the following optimization problem:
\begin{equation}\label{eq:beta_solve}
    \textstyle \beta \gets \argmax_{\beta} \left\{ \epsilon'(\alpha) \leq \frac{\Tilde{\epsilon}}{n_{\text{test}} T_{\text{max}}} \right\}
\end{equation}
where $\Tilde{\epsilon}$ is the converted $(\Tilde{\epsilon}, \alpha)$-RDP from $(\epsilon, \delta)$-DP using Theorem \ref{thm:rdp-dp}. The idea is that we want to maximize the radius $\beta\alpha$ by increasing $\beta$ while satisfying the constraint that the subsampled privacy loss $\epsilon'(\alpha)$ is at most $\frac{\Tilde{\epsilon}}{n_{\text{test}} T_{\text{max}}}$. Just like $\lambda$, we use the Bisection method to numerically solve this optimization problem. 

Next, to get an upper bound on the total privacy loss over the entire inference, we calculate the total number of queries that will be performed by our algorithm, which is $n_{\text{test}}T_{\max}$ where $T_{\max}$ is the max generation length of each response $\mathbf{y}$ and $n_{\text{test}}$ is the number of responses that the model will answer. Hence, the total privacy loss is $(\alpha, n_{\text{test}} T_{\max}\epsilon'(\alpha))$-RDP by the composition theorem \ref{thm:composition}. By Equation \ref{eq:beta_solve}, this simplifies to $(\alpha, \Tilde{\epsilon})$-RDP, which converts back to its $(\epsilon, \delta)$-DP counterpart. 

\clearpage

\section{Additional Details and Pseudocode for \dpsmozo{+On} and \dpsmozo{+Off}}\label{app:add_algs}
For completeness, we provide full details on \dpsmozo{+On} and \dpsmozo{+Off} by including the complete corresponding pseudocode. \cref{alg:dpsmozo_on} shows the pseudocode for \dpsmozo{+On} and \cref{alg:dpsmozo_off} shows the pseudocode for \dpsmozo{+Off}.

\begin{algorithm}[H]
    \caption{\dpsmozo{+On}: Applying \dpsmozo{} at Online Phase}
    \label{alg:dpsmozo_on}
    \textbf{Input:} An LLM $\theta$, a set of input-output examples $D$, a set of input queries $D_{\text{test}}$, number of input queries $n_{\text{test}}$, number of top public indices $k$, Renyi order $\alpha$, max generation size $T_{\max}$, number of shots $n_{\text{shots}}$, Target privacy $\epsilon$, Subsampled privacy loss $\epsilon'(\alpha)$ \\ 
    \textbf{Output:} output queries $\mathbf{y}_1, ..., \mathbf{y}_{|D_{\text{test}}|}$
    \begin{algorithmic}[1]
        \STATE \textbf{Return} $\texttt{Online}(D)$
        \FUNCTION{$\texttt{Online}(D)$}
        \STATE $\mathbf{Y} \gets []$
        \STATE Convert to target Renyi Privacy $\Tilde{\epsilon}$ using \cref{thm:rdp-dp}
        \STATE $\beta \gets \argmax_{\beta} \left\{\epsilon'(\alpha) \leq \frac{\Tilde{\epsilon}}{n_{\text{test}} T_{\max}} \right\}$
        \FOR{$\mathbf{x}\in D_{\text{test}}$}
            \STATE $\mathbf{y} \gets \dpsmozo{}(\beta, \alpha, \theta, D, \mathbf{x}, k, T_{\max}, n_{\text{shots}})$
            \STATE $\mathbf{Y} \gets \mathbf{Y} + [\mathbf{y}]$
        \ENDFOR
    \STATE \textbf{Return} $\mathbf{Y}$
    \ENDFUNCTION
    \end{algorithmic}
\end{algorithm}
\begin{algorithm}[H]
    \caption{\dpsmozo{+Off}: Applying \dpsmozo{} at Offline Phase}
    \label{alg:dpsmozo_off}
    \textbf{Input:} An LLM $\theta$, a set of input-output examples $D$, a set of public inputs $D_{\text{pub}}$, a set of input queries $D_{\text{test}}$, number of input queries $n_{\text{test}}$, number of top public indices $k$, Renyi order $\alpha$, Synthetic max generation size $\tilde{T}_{\max}$, query output max generation size $T_{\max}$, number of shots $n_{\text{shots}}$, Target privacy $\epsilon$, Subsampled privacy loss $\epsilon'(\alpha)$ \\ 
    \textbf{Output:} output queries $\mathbf{y}_1, ..., \mathbf{y}_{|n_{\text{test}}|}$
    \begin{algorithmic}[1]
        \STATE $D_{\text{syn}} \gets \texttt{Offline}(D, D_{\text{pub}})$
        \STATE \textbf{Return} $\texttt{Online}(D_{\text{syn}})$
        \FUNCTION{\texttt{Offline}$(D, D_{\text{pub}})$}
        \STATE $D_{\text{syn}} \gets []$
        \STATE Convert to target Renyi Privacy $\Tilde{\epsilon}$ using \cref{thm:rdp-dp}
        \STATE $\beta \gets \argmax_{\beta} \left\{\epsilon'(\alpha) \leq \frac{\Tilde{\epsilon}}{n_{\text{test}} \Tilde{T}_{\max}} \right\}$
        \FOR{$\mathbf{x}\in D_{\text{pub}}$}
            \STATE $\mathbf{y} \gets \dpsmozo{}(\beta, \alpha, \theta, D, \mathbf{x}, k, \Tilde{T}_{\max}, n_{\text{shots}})$
            \STATE $D_{\text{syn}} \gets D_{\text{syn}} + [(\mathbf{x}, \mathbf{y})]$
        \ENDFOR
    \STATE \textbf{Return} $D_{\text{syn}}$
    \ENDFUNCTION
    \FUNCTION{\texttt{Online}$(D_{\text{syn}})$}
    \STATE $\mathbf{Y} \gets [] $
    \FOR{$\mathbf{x}\in D_{\text{test}}$}
        \STATE \textcolor{blue}{// Using standard non-private decoding function}
        \STATE $\mathbf{y} \gets \texttt{Decode}(\theta, D_{\text{syn}}, \mathbf{x}, T_{\max}, n_{\text{shots}})$
        \STATE $\mathbf{Y} \gets \mathbf{Y} + [\mathbf{y}]$
    \ENDFOR
    \STATE \textbf{Return} $\mathbf{Y}$
    \ENDFUNCTION
    \end{algorithmic}
\end{algorithm}

\section{More Details on Experimental Setup}\label{app:add_exp_setup}
\citet{wu2023privacy} proposed three private aggregation methods for language generation. We chose the Embedding Space Aggregation method because (1) it is the least-involved method to reimplement, and (2) it is the second-best-performing method for dialog summarization, performing close to the best-performing method. The Embedding method from \citet{wu2023privacy} can be found in Algorithm \ref{alg:baseline}. This method requires obtaining embedding vectors for output sentences, so we use \texttt{UAE-Large-V1}, a recent pre-trained sentence embedding model from \citet{li2023angle} that has demonstrated strong performance in various sentence embedding benchmarks. In our experiments, we follow the same hyperparameter values as the experimental evaluations in \citet{wu2023privacy} by setting $N=100$, $C=100$, and $|D_i|=4$. Calculating $\sigma$ follows from \citet{wu2023privacy} by using the \texttt{priv_accountant}\footnote{\url{https://github.com/microsoft/prv_accountant}} package. 

\cref{fig:e2e_prompts}, \cref{fig:spider_prompts}, and \cref{fig:samsum_prompts} display example one-shot and zero-shot prompts used for each dataset. \cref{tab:dataset_stats_table} displays the dataset statistics used in the experimental evaluations.

\newcommand{\zeroshot}{O_C}

\begin{algorithm}[h!]
\caption{Differentially Private In-Context Learning via Embedding Space Aggregation}
\label{alg:dpemb}
\begin{algorithmic}[1]
\INPUT{Private data $D$, query $\mathbf{x}$, model $p_{\theta}$, noise $\sigma$, number of subsets $N$,
public candidate sentences obtained by zero-shot predictions $\zeroshot$ with total number of $C$}
\STATE \textbf{Partition} $D_1, D_2, \dots, D_N \leftarrow D$. 
\FOR{$i \in \{1, \dots, N\}$}
    \STATE Form exemplar-query pair $D_i^{\mathbf{x}} = D_i \cup \{\mathbf{x}\}$.
    \STATE Obtain model output sentence $O_i(\mathbf{x}) = p_\theta(D_i^{\mathbf{x}})$.
    \STATE Project $O_i(\mathbf{x})$ into embedding vector $E_i(\mathbf{x})$. 
\ENDFOR
\STATE Take the mean of all embedding vectors $E = \frac{1}{N} \sum_{i=1}^N E_i$.
\STATE Adding noise and obtain the noisy embedding  $\hat E = E + \mathcal{N}\left(0, \sigma^2 I \right)$. 
\STATE \textbf{Return} the sentence $\argmax_{o \in \zeroshot} \textbf{CosineSimilarity}(o, \hat{E})$.
\end{algorithmic}\label{alg:baseline}
\end{algorithm}
\begin{figure}[h!]
    \centering
    \begin{subfigure}{0.49\textwidth}
        \begin{mdframed}[backgroundcolor=gray!5, linecolor=black, linewidth=1pt]
        \begin{center}
            \textbf{One-shot Prompt} 
        \end{center}
        
        \noindent Please convert the structured data into natural language. \\
        \textcolor{red}{\textbf{Input:} \\
        name[The Vaults], eatType[pub], priceRange[more than £30], customer rating[5 out of 5], near[Café Adriatic] \\
        \textbf{Answer:} The Vaults pub near Café Adriatic has a 5 star rating. Prices start at £30.} 
        \\
        
        \noindent \textcolor{blue}{\textbf{Input:} \\
        name[Blue Spice], eatType[coffee shop], area[city centre] \\
        \textbf{Answer:} \\
        }
        \end{mdframed}
    \end{subfigure}
    \begin{subfigure}{0.49\textwidth}
        \begin{mdframed}[backgroundcolor=gray!5, linecolor=black, linewidth=1pt]
        \begin{center}
            \textbf{Zero-shot} 
        \end{center}
        
        \noindent Please convert the structured data into natural language. \\
        \noindent \textcolor{blue}{\textbf{Input:} \\
        name[Blue Spice], eatType[coffee shop], area[city centre] \\
        \textbf{Answer:} \\
        }
        \end{mdframed}
    \end{subfigure}
    \caption{Example of one-shot and zero-shot prompts used for the E2E dataset where \textcolor{red}{red text} is an example input-output pair $(\mathbf{x}_i, \mathbf{y}_i)$ and \textcolor{blue}{blue text} is the input query $\boldsymbol{x}$.}
    \label{fig:e2e_prompts}
\end{figure}
\begin{figure}[h!]
    \centering
    \begin{subfigure}{0.49\textwidth}
        \begin{mdframed}[backgroundcolor=gray!5, linecolor=black, linewidth=1pt]
        \begin{center}
            \textbf{One-shot Prompt} 
        \end{center}
        
        \noindent Please make the sentences easier to read and understand. \\
        \textcolor{red}{\textbf{Input:} \\
        lemon meringue pie is a type of baked pie usually served for dessert typically prepared with a bottom pie crust but with no upper crust.  \\
        \textbf{Output:} \\ lemon meringue pie is a type of sweet pie and dessert. }
        \\
         
        \noindent \textcolor{blue}{\textbf{Input:} \\
        for example king bhumibol was born on monday so on his birthday throughout thailand will be decorated with yellow color. \\
        \textbf{Answer:} \\
        }
        \end{mdframed}
    \end{subfigure}
    \begin{subfigure}{0.49\textwidth}
        \begin{mdframed}[backgroundcolor=gray!5, linecolor=black, linewidth=1pt]
        \begin{center}
            \textbf{Zero-shot} 
        \end{center}
        
        \noindent Make the input a little simpler. \\
        \noindent \textcolor{blue}{\textbf{Input:} \\
        for example king bhumibol was born on monday so on his birthday throughout thailand will be decorated with yellow color. \\
        \textbf{Answer:} \\
        }
        \end{mdframed}
    \end{subfigure}
    \caption{Example of one-shot and zero-shot prompts used for the WikiLarge dataset where \textcolor{red}{red text} is an example input-output pair $(\mathbf{x}_i, \mathbf{y}_i)$ and \textcolor{blue}{blue text} is the input query $\boldsymbol{x}$.}
    \label{fig:spider_prompts}
\end{figure}
\begin{figure}[h!]
    \centering
    \begin{subfigure}{0.49\textwidth}
        \begin{mdframed}[backgroundcolor=gray!5, linecolor=black, linewidth=1pt]
        \begin{center}
            \textbf{One-shot Prompt} 
        \end{center}
        
        \noindent \textcolor{red}{\textbf{Dialogue:} \\
        Mary: Hi Mike!\\ 
        Mike: Hello :)\\
        Mary: do u have any plans for tonight?\\
        Mike: I'm going to visit my grandma.\\
        Mike: You can go with me.\\
        Mike: She likes u very much.\\
        Mary: Good idea, i'll buy some chocolate for her.\\
        \textbf{Summarize the above dialogue:} Mike and Mary are going to visit Mike's grandma tonight. Mary will buy her some chocolate.} 
        \\
        
        \noindent \textcolor{blue}{\textbf{Dialogue:} \\
        Josh: Stephen, I think you've accidentaly taken my notebook home\\
        Stephen: wait lemme check\\
        Stephen: nope, I don't see it anywhere\\
        Jack: oh shit, I've got it xDDD I don't even know why\\
        Josh: xDDD ok, no problem, cool I know where it is\\
        Jack: I'll bring it tomorow\\
        \textbf{Summarize the above dialogue:} }
        \end{mdframed}
    \end{subfigure}
    \begin{subfigure}{0.49\textwidth}
        \begin{mdframed}[backgroundcolor=gray!5, linecolor=black, linewidth=1pt]
        \begin{center}
            \textbf{Zero-shot} 
        \end{center}
        
        \noindent \textcolor{red}{} 
        \noindent \textcolor{blue}{\textbf{Dialogue:} \\
        Josh: Stephen, I think you've accidentaly taken my notebook home\\
        Stephen: wait lemme check\\
        Stephen: nope, I don't see it anywhere\\
        Jack: oh shit, I've got it xDDD I don't even know why\\
        Josh: xDDD ok, no problem, cool I know where it is\\
        Jack: I'll bring it tomorow\\
        \textbf{Summarize the above dialogue:}}
        \end{mdframed}
    \end{subfigure}
    \caption{Example of one-shot and zero-shot prompts used for the SAMSum dataset where \textcolor{red}{red text} is an example input-output pair $(\mathbf{x}_i, \mathbf{y}_i)$ and \textcolor{blue}{blue text} is the input query $\boldsymbol{x}$.}
    \label{fig:samsum_prompts}
\end{figure}
\begin{table}[H]
\caption{Dataset statistics used for experimental evaluations.}
\centering
\begin{tabular}{ccccc}
\toprule
Task & \#Train & \#Validation & \#Test & Task description \\ \toprule
SAMSum & 14,732 & 818 & 819 & Dialogue summarization \\ \midrule
E2E & 42,061 & 4,672 & 4693 & Data-to-text generation \\ \midrule
WikiLarge & 149,000 & 494 & 191 & Text-to-data generation  \\ \midrule
\end{tabular}
\label{tab:dataset_stats_table}
\end{table}

\cref{tab:hyperparam_table} displays the hyperparameter table. \citet{wu2023privacy} uses concat-based decoding and the subsample-and-aggregate paradigm, while \dpsmozo{} follows ensemble-based decoding. Hence, both methodologies utilize ensembles/subsets, so we introduce two hyperparameters, \#Subsets, the number of subsets, and Subset Size, the size of each subset, to help unify the notation and hyperparameters. For example, $n_{\text{shots}}=4$ is equivalent to \#Subets=4 and Subset Size=1 for ensemble-based decoding. Note that we also needed to introduce an additional parameter, $\Tilde{T}_{\max}$ which is the max generation size for the synthetic example output, while $T_{\max}$ is the max generation size for query output.

\begin{table}[H]
\caption{ICL hyperparameters used for experimental evaluations.}
\centering
\begin{tabular}{l|cccccc}
\toprule
Dataset & Method & \#Subsets & Subset Size & $T_{\max}$ & $\Tilde{T}_{\max}$ & $k$ \\ \toprule
\multirow{4}{*}{SAMSum} & \citet{wu2023privacy} & 100 & 4 & 50 & - & 100 \\ \cmidrule(lr){2-7}
& \dpsmozo{+On} & 4 & 1 & 50 & - & 100 \\ \cmidrule(lr){2-7}
& \dpsmozo{+Off} & 4 & 1 & 50 & 30 & 100 \\ \midrule
\multirow{4}{*}{E2E} & \citet{wu2023privacy} & 100 & 4 & 25 & - & 100 \\ \cmidrule(lr){2-7}
& \dpsmozo{+On} & 4 & 1 & 25 & - & 100 \\ \cmidrule(lr){2-7}
& \dpsmozo{+Off} & 4 & 1 & 25 & 20 & 100 \\ \midrule
\multirow{4}{*}{WikiLarge} & \citet{wu2023privacy} & 100 & 4 & 25 & - & 100 \\ \cmidrule(lr){2-7}
& \dpsmozo{+On} & 4 & 1 & 25 & - & 100 \\ \cmidrule(lr){2-7}
& \dpsmozo{+Off} & 4 & 1 & 25 & 20 & 100 \\ 
\bottomrule
\end{tabular}
\label{tab:hyperparam_table}
\end{table}

Additionally, we list the privacy parameters used by \dpsmozo{} for each dataset in \cref{tab:priv_param_table}. This includes approximate DP $\epsilon$, converted RDP $\tilde{\epsilon}$, Renyi Divergence order $\alpha$, max number of queries $T_{\max}n_{{\text{test}}}$, and target leakage $\beta$.

\begin{table}[H]
\caption{Privacy parameters used in experimental evaluations for \dpsmozo{}.}
\centering
\begin{tabular}{l|cccccc}
\toprule
Dataset & $\epsilon$ & $\tilde{\epsilon}$ & $\alpha$ & $T_{\max}n_{{\text{test}}}$ &$\beta$ \\ \toprule
\multirow{4}{*}{SAMSum} & 1.0 & 0.539 & 14 & 5000 & 0.081 \\ \cmidrule(lr){2-6}
& 2.0 & 1.059 & 8 & 5000 & 0.179 \\ \cmidrule(lr){2-6}
& 4.0 & 2.226 & 5 & 5000 & 0.342 \\ \midrule
\multirow{4}{*}{E2E} & 1.0 & 0.502 & 15 & 2500 & 0.115 \\ \cmidrule(lr){2-6}
& 2.0 & 1.062 & 9 & 2500 & 0.220 \\ \cmidrule(lr){2-6}
& 4.0 & 2.377 & 6 & 2500 & 0.370 \\ \midrule
\multirow{4}{*}{WikiLarge} & 1.0 & 0.527 & 18 & 2500 & 0.120 \\ \cmidrule(lr){2-6}
& 2.0 & 1.038 & 10 & 2500 & 0.242 \\ \cmidrule(lr){2-6}
& 4.0 & 2.159 & 6 & 2500 & 0.445 \\
\bottomrule
\end{tabular}
\label{tab:priv_param_table}
\end{table}

\subsection{Experimental Setup for MIAs}
\label{app:mia_setup}

To rigorously evaluate the empirical privacy leakage through Membership Inference Attacks (MIAs), we detail the experimental setup and procedures as follows:

\subsubsection{Dataset and Test Selection}
Following \citet{duan2024flocks}, for each experiment, we select 51 examples from the training set to form the test set, denoted as $D_{\text{test}} = \{\mathbf{x}_i\}_{i=1}^{51}$. Then, in each attack, we select one $\mathbf{x}'\in D_{\text{test}}$ as the member example and the rest 50 examples as nonmember examples to create a distribution between members and non-members (1 vs 50) that reflects the realistic scenario where only a small proportion of the candidate data targeted by the adversary are members \citet{duan2024flocks}.

\subsubsection{Membership Score Computation}
For a designated member example $(\mathbf{x}', \mathbf{y}')$, we compute its membership score as:
\begin{equation}
p_\theta(\mathbf{y}'| \mathbf{x}', \mathbf{y}', \mathbf{x}')
\end{equation}
where $\mathbf{x}'$ serves as both the query and demonstration example.
For each non-member example $\mathbf{x}i \in D{\text{test}} \setminus {\mathbf{x}'}$, we compute its score as:
\begin{equation}
p_\theta(\mathbf{y}_i| \mathbf{x}', \mathbf{y}', \mathbf{x}_i)
\end{equation}
where $\mathbf{x}'$ remains as the demonstration example.
\subsubsection{Attack Procedure}
The attack process is structured as follows:
\begin{enumerate}
\item Perform 51 individual attacks, designating a different example from $D_{\text{test}}$ as the member example $(\mathbf{x}', \mathbf{y}')$ for each attack. This step will ensure the same level of difficulty for member and non-member on average. Otherwise if members were slightly easier or harder to answer correctly than non-members, it could introduce variability in the AUC-ROC score commensurable to the actual privacy leakage effect of the algorithm when the AUC-ROC score is close to 0.5.

\item For each attack, compute 51 scores: one for the designated member example and 50 for non-member examples.
\item Aggregate these scores to form a binary classification problem for distinguishing members from non-members.
\item Repeat the entire experiment 5 times, using different random selections of $D_{\text{test}}$ in each repetition.
\item Average the results across all repetitions to ensure robustness and minimize variability in the evaluation.
\end{enumerate}

\subsubsection{Evaluation Metric}
To quantify the success of the membership inference attack, we use the Area Under the Receiver Operating Characteristic Curve (AUC-ROC). A score of 50\% indicates that the adversary performs no better than random guessing, while a higher score suggests increased privacy leakage.

\end{document}